\documentclass{article}

\usepackage{arxiv}

\usepackage[utf8]{inputenc} 
\usepackage[T1]{fontenc}    
\usepackage{hyperref}       
\usepackage{url}            
\usepackage{booktabs}       
\usepackage{amsfonts}       
\usepackage{nicefrac}       
\usepackage{microtype}      
\usepackage{lipsum}		
\usepackage{graphicx}
\usepackage{natbib}
\usepackage{doi}

\usepackage{amssymb}
\usepackage{amsmath}
\usepackage{latexsym}
\usepackage{nicefrac,xfrac}

\usepackage{url}
\usepackage{xcolor}
\usepackage{hyperref}

\usepackage{verbatim}

\usepackage{placeins}

\newcommand{\xb}{\textbf{x}} 
\newcommand{\yb}{\textbf{y}} 
\newcommand{\revisiontwo}[1]{{\color{black}#1}}

\newcommand{\figBatches}[1]{
\includegraphics[width=0.11\linewidth]{images/Example_Entropy/#1}}

\newcommand{\figSeg}[1]{
\includegraphics[height=0.11\linewidth]{images/Segmentation/#1}}

\newcommand{\ppm}[1]{\footnotesize($\pm$#1)}

\DeclareMathOperator*{\argmax}{argmax} 

\usepackage{algpseudocode}
\usepackage{algorithm}  

\usepackage{float}
\usepackage{caption}
\usepackage{subcaption}
\usepackage{dblfloatfix}    

\usepackage{booktabs}
\usepackage{tabularx}
\usepackage{array}
\usepackage{multirow}
\usepackage{arydshln} 
\usepackage{adjustbox} 
\usepackage{tabularx} 

\title{Active learning for medical image segmentation with stochastic batches}

\date{} 					

\author{M\'elanie Gaillochet \\
	\'ETS Montr\'eal\\
	\And
	Christian Desrosiers \\
        \'ETS Montr\'eal\\
        \And
        Herv\'e Lombaert \\
        \'ETS Montr\'eal
        }
        
\renewcommand{\undertitle}{}
\renewcommand{\shorttitle}{Active learning for medical image segmentation with stochastic batches}

\hypersetup{
pdftitle={A template for the arxiv style},
pdfsubject={q-bio.NC, q-bio.QM},
pdfauthor={David S.~Hippocampus, Elias D.~Striatum},
pdfkeywords={First keyword, Second keyword, More},
}

\begin{document}
\maketitle

\begin{abstract}
The performance of learning-based algorithms improves with the amount of labelled data used for training. Yet, manually annotating data is particularly difficult for medical image segmentation tasks because of the limited expert availability and intensive manual effort required. To reduce manual labelling, active learning (AL) targets the most informative samples from the unlabelled set to annotate and add to the labelled training set. On the one hand, most active learning works have focused on the classification or limited segmentation of natural images, despite active learning being highly desirable in the difficult task of medical image segmentation. On the other hand, uncertainty-based AL approaches notoriously offer sub-optimal batch-query strategies, while diversity-based methods tend to be computationally expensive. Over and above methodological hurdles, random sampling has proven an extremely difficult baseline to outperform when varying learning and sampling conditions. This work aims to take advantage of the diversity and speed offered by random sampling to improve the selection of uncertainty-based AL methods for segmenting medical images. More specifically, we propose to compute uncertainty at the level of batches instead of samples through an original use of stochastic batches (SB) during sampling in AL. \revisiontwo{Stochastic batch querying is a simple and effective add-on that can be used on top of any uncertainty-based metric. Extensive experiments} on two medical image segmentation datasets show that \revisiontwo{our strategy} consistently improves conventional uncertainty-based sampling methods. Our method can hence act as a strong baseline for medical image segmentation. \revisiontwo{The code is available on: \href{https://github.com/Minimel/StochasticBatchAL.git}{https://github.com/Minimel/StochasticBatchAL.git}}.
\end{abstract}


\section{Introduction}
\label{sec:intro}
Data annotation is fundamental to medical imaging. Notably, the performance of segmentation algorithms depends on the amount of annotated training data. The manual annotation of pixel-level ground truth is therefore highly sought but remains difficult to obtain due to two challenging problems. First, the pixel-wise annotation of entire biological structures is a laborious and expensive task that requires highly trained clinicians. Second, image acquisition grows faster than the experts' ability to manually process the data, leaving large datasets mostly unlabelled. Clinicians can realistically annotate only small sets of images with a limited capacity to scale up. This constraint creates a need for strategies that reduce the crucial but arduous annotation efforts in medical imaging.

To maximize the performance of a model with reduced annotated data during training, two types of approaches can unleash the potential of unlabelled data: active learning and semi-supervised learning. Active learning (AL) aims to identify the best samples to annotate and use during training. Meanwhile, semi-supervised learning seeks to improve the representation learned from data by exploiting unlabelled samples in addition to the few labelled ones. However, this approach still leaves the question of choosing which samples to use for the labelled set, underlining the importance of active learning.

Images in the training set do not contribute equally to the performance of learning-based algorithms \citep{settles_active_2009}. Given a large unlabelled dataset, active learning 
overcomes labelled data \revisiontwo{scarcity} by incrementally identifying the most valuable samples to be annotated and added to a training set \citep{budd_survey_2021,ren_survey_2021}. Actively selecting which data to label conceivably maximizes the performance of machine learning models with a minimum amount of labelled data. AL strategies also have the potential of accelerating training convergence and improving robustness by targeting specific types of data points \citep{nath_diminishing_2021}. 


Active learning methods can be divided into three broad categories: uncertainty-based sampling strategies, representative-based sampling strategies and hybrid approaches \citep{settles_active_2009,budd_survey_2021}. 
Uncertainty-based methods assume that the most valuable samples to annotate are the ones for which the current model is least confident. These methods, which differ in ways of calculating uncertainty, are however susceptible to target outlier samples or redundant information, particularly when querying batches of samples. 
To avoid bias towards narrow locals in distributions, representative-based and hybrid approaches try to diversify the set of candidate samples. Ensuring such diversity generally relies on learning a latent data representation, which requires estimating pairwise distances between all samples or computing their marginal distribution. These strategies consequently hardly scale satisfyingly to high dimensions. Consequently, the majority of active learning approaches applied to computer vision focus on lower-dimensional tasks such \revisiontwo{as} classification, while AL approaches for segmentation tend to focus on natural images with several thousands of annotated images \citep{sinha_variational_2019,huang_semi-supervised_2021,kim_task-aware_2021,xie_towards_2022}. Due to its high-dimensional nature, medical image segmentation remains \revisiontwo{an ongoing challenge in} active learning, despite the substantial need to minimize the high cost of manual annotation from clinical expertise.

A limited yet increasing number of works acknowledges that random sampling is, in practice, a painstakingly difficult baseline to outperform in active learning \citep{kirsch_batchbald_2019,mittal_parting_2019,nath_diminishing_2021,munjal_towards_2022,burmeister_less_2022}. Indeed, the gains of AL strategies over random sampling are often inconsistent across different experimental setups. 
For example, varying the sampling budget can cancel the improvements originally observed for such strategies \citep{bengar_reducing_2021,munjal_towards_2022}. Similarly, existing methods for AL tend to be sensitive to the model architecture, hyperparameters and regularization used during training \citep{mittal_parting_2019,munjal_towards_2022}. These hurdles hinder AL advances in medical image segmentation.

This paper intends to address the limitations of current AL methods, notably their drawback of selecting batches solely based on per-sample uncertainty, the computational cost of ensuring diversity, \revisiontwo{and the significantly varying amounts of robustness in performance across experimental setups}. Our work proposes to leverage the power of randomness during uncertainty-based batch sampling to improve the overall segmentation performance of AL models. 

\begin{figure*}[!t]
    \centering
    \includegraphics[width=\linewidth]{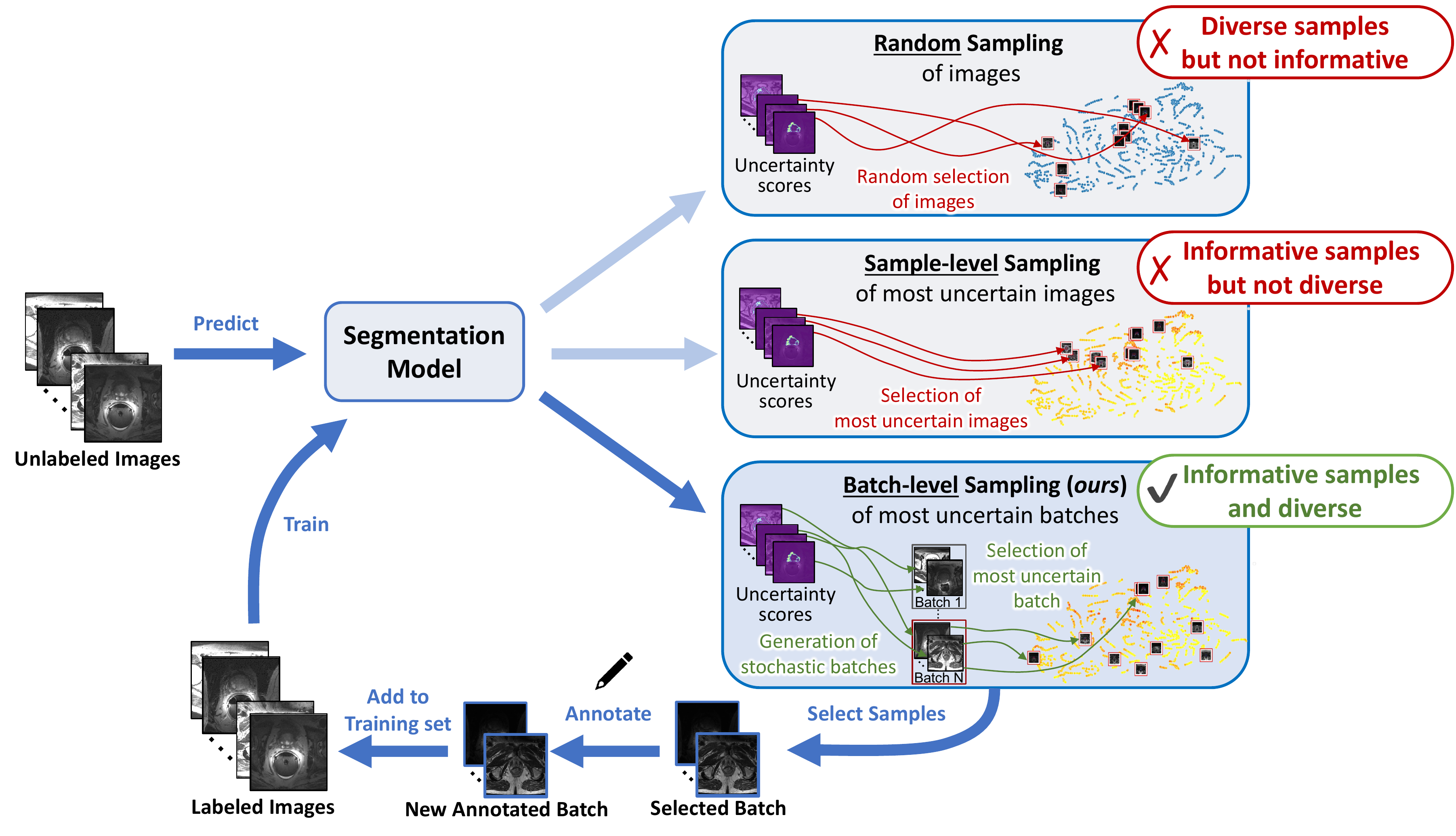}
     \caption{\revisiontwo{\textbf{Stochastic batch AL for uncertainty-based sampling}. Our sampling method combines the diversity of random sampling with the informativeness of uncertainty-based sampling. Adding our stochastic batch paradigm enables the data uncertainty to be estimated in a broader \textit{batch-level} selection rather than a \textit{sample-level} selection. After selecting a candidate set of unlabelled samples, the set is annotated and added to the existing labelled set. Finally, the segmentation model is retrained.}}
    \label{fig:stochastic_batches}
\end{figure*}

\subsection*{Contributions}
\label{subsec:contribution}

We introduce the use of \revisiontwo{stochastic batch (SB) querying, a simple and effective add-on to uncertainty-based AL strategies, compatible with any uncertainty metric} (see Fig.\ref{fig:stochastic_batches}). Our stochastic batch sampling strategy proves advantageous by:
\begin{enumerate}
    \item minimizing the problem of uncertainty-based strategies, often susceptible to query samples with redundant information;
    \item allowing uncertainty-based AL strategies to benefit from a larger diversity of samples in a simple and computationally-efficient way; and
    \item \revisiontwo{providing noticeably consistent gains across different experimental settings, as shown by our extensive ablation studies.}
\end{enumerate}


\section{Literature review}
\label{sec:literature_review}
Active learning methods maximize the future model performance by augmenting the current labelled training set with the most informative unlabelled samples. AL approaches mainly fall into uncertainty-based, representative-based or hybrid strategies, each described next.

\subsection{Uncertainty-based AL methods}
Uncertainty is one of the most prevalent criteria for sampling in active learning. Uncertainty-based methods query samples for which the current model is least confident \citep{settles_active_2009}. 
AL strategies for deep learning-based models have initially applied traditional AL methods that identify difficult examples using simple heuristics.
However, in practice, they still hardly scale to high-dimensional data \citep{beluch_power_2018} or are not consistently effective for deep learning models that rely on batch selection \citep{sener_active_2018,ren_survey_2021}. Hence, subsequent work has combined traditional uncertainty measures, such as the entropy of the output probabilities, with measures of geometric uncertainty \citep{konyushkova_geometry_2019} or with the pseudo-labelling of samples with confident predictions \citep{wang_cost-effective_2017}. Similarly, \cite{gal_deep_2017} and \cite{kirsch_batchbald_2019} \revisiontwo{adapt} existing heuristics to a Bayesian framework through Monte Carlo dropout. More recently, \cite{yoo_learning_2019} developed a new uncertainty measure based on the predicted loss from the intermediate representations of the model. 
Although widely popular, purely uncertainty-based strategies relying on batch selection are susceptible to query samples with redundant information. However, manually annotating similar samples is a waste of annotation resources. Moreover, incorporating a set of similar samples to the labelled training set could bias the model towards an area outside the true data distribution. These samples could hence hamper rather than improve model generalization.

\subsection{Representative-based AL methods}
As opposed to uncertainty-based approaches, representative-based AL methods aim at diversifying the batch of candidate samples to improve the future performance of the model \citep{settles_active_2009}. One of the main representative-based approaches, \revisiontwo{Core-set} \citep{sener_active_2018}, identifies the most diverse and representative samples by minimizing the distance between \revisiontwo{the latent representations of labelled and unlabelled images, as given by the task model}. \revisiontwo{Core-set} aims for the model to perform as well with the candidate set as it would with the entire dataset. While specifically designed to be applied to complex models such as Convolutional Neural Networks (CNNs), \revisiontwo{core-set} selection does not scale well to high-dimensional data since it requires computing the Euclidean distance between all pairs of data samples. A later work, VAAL \citep{sinha_variational_2019}, learns a smooth latent-state representation of the input data via a variational auto-encoder (VAE). VAAL then selects samples different from the ones already labelled based on the learnt latent representation. Since the VAE is task-agnostic, VAAL can, however, easily query outlier data. In addition, it provides no mechanism to avoid choosing overlapping samples and requires \revisiontwo{careful} tuning of its added modules. 

\subsection{Hybrid AL strategies}
Against the limitations of uncertainty-based methods, hybrid strategies try to find a balance between uncertainty and diversity measures to identify the most informative samples \citep{settles_active_2009}. They usually combine existing approaches. \revisiontwo{An} early study proposed to adaptively choose the best AL strategies from a candidate set of methods \citep{hsu_active_2015}. \revisiontwo{However, most hybrid methods} first compute model uncertainty before ensuring sample diversity through a similarity metric. For instance, Suggestive Annotation \citep{yang_suggestive_2017} applies \revisiontwo{core-set} selection on a subgroup of the most uncertain samples obtained through bootstrapping. BADGE \citep{ash_deep_2020} uses gradient embeddings to account for uncertainty (uncertain samples will have a gradient embedding with higher norm) and employs Kmeans++ initialization on top of these embeddings to ensure the diversity of selected samples. \cite{nath_diminishing_2021} \revisiontwo{combine} prevailing mutual information and entropy measures to ensure diversity and optimize training by duplicating difficult samples. Observing that uncertainty-based approaches fail to exploit the data distribution and representative-based approaches are task-agnostic, Task-aware VAAL \citep{kim_task-aware_2021} incorporates the uncertainty measure proposed by the method Learning Loss \citep{yoo_learning_2019} to VAAL's \citep{sinha_variational_2019} latent representation. While these studies rely on a two-step approach, \cite{sourati_intelligent_2019} directly \revisiontwo{solve} an optimization problem for batch-mode sampling, yielding a distribution of candidate samples rather than specific examples. However, just like representative-based AL strategies, most of these works are difficult to scale due to their computational complexity \citep{ ash_deep_2020,nath_diminishing_2021,sourati_intelligent_2019,yang_suggestive_2017}.  
Alternatively, they may require external modules, which increase the range of parameters to tune and learn \citep{kim_task-aware_2021}.

\subsection{AL for medical image segmentation}
High-dimensional data remains a particularly challenging problem in AL \citep{ren_survey_2021}. Therefore, most studies on AL applied to computer vision primarily focus on low-dimensional annotation tasks such as image classification \citep{gal_deep_2017,wang_cost-effective_2017,sener_active_2018,beluch_power_2018,sourati_intelligent_2019,gao_consistency-based_2020,ash_deep_2020,zhang_boostmis_2022}. Moreover, approaches tackling pixel-wise annotations predominantly address the segmentation of natural images \citep{sinha_variational_2019,huang_semi-supervised_2021,kim_task-aware_2021,xie_towards_2022}.

Earlier work applying AL to medical image segmentation has relied on geometric priors to query planes or supervoxels of maximum uncertainty, without adopting deep learning-based models \citep{top_active_2011,konyushkova_introducing_2015,konyushkova_geometry_2019}. One of the initial deep AL frameworks for this task, Suggestive Annotation \citep{yang_suggestive_2017}, uses bootstrapping to estimate sample uncertainty and a greedy cosine similarity measure to evaluate the similarity between the candidate set and the unlabelled pool. Similarly, \cite{li_attention_2020} \revisiontwo{propose to select a candidate set with a high disagreement among the predictions of K models and a minimal discrepancy between the labelled and unlabelled sets}. Instead of relying on multiple models, \cite{ozdemir_active_2018} employ a Bayesian network with Monte Carlo dropout to compute prediction variance, and adopt a Borda-count-based sampling strategy to find the best-ranked candidates in terms of uncertainty and representativeness. An extension of this approach instead computes the representativeness with an infoVAE \citep{zhao_infovae_2019} for a maximum-likelihood sampling in the latent space \citep{ozdemir_active_2021}. \cite{nath_diminishing_2021} build a mutual information-based metric, computed between the labelled and unlabelled pools, to ensure the diversity of the candidate set. However, these approaches tend to be computationally expensive and challenging to scale to large datasets. Instead of relying on a 2-step approach, \cite{stoyanov_active_2018} propose a method based on the Fisher information to directly solve an optimization problem that outputs a distribution to sample from.
\revisiontwo{Alternative} approaches have opted for membership query synthesis as an AL strategy, producing synthetic samples for annotation. For instance, \cite{mahapatra_efficient_2018} employ a conditional generative adversarial network (cGAN) to generate realistic-looking chest X-ray images conditioned on real images, \revisiontwo{and} a Bayesian neural network to select which ones would be most informative when used as training data. 
\revisiontwo{Other approaches propose a sample selection strategy which also covers the initial labelled set \citep{smailagic_medal_2018,nath_warm_2022,li_hal-ia_2023}}. 
Recently, a comparative study of existing strategies for 3D medical image segmentation found that random sampling and strided sampling served as particularly strong baselines for this type of task  \citep{burmeister_less_2022}. The study also observed that representative-based strategies did not perform well in early stages, which the authors attribute to poor feature vectors generated by the model trained on very few labelled samples.

\section{Methods}
\label{sec:method}
Given a labelled set $\mathcal{D}_L = \{ (\xb^{(j)}, \yb^{(j)}) \}_{j = 1}^{N}$, with data $\xb \in \mathbb{R}^{H \times W}$ and segmentation mask $\yb \in \mathbb{R}^{C \times H \times W}$ ($H$ and $W$ are respectively the image height and width, and $C$ is the number of classes), we train a fully-supervised segmentation model $f_{\theta}(\cdot)$ parameterized by $\theta$ with labelled samples from $\mathcal{D}_L$.

After training the model $f_{\theta}$ with $\mathcal{D}_L$ (corresponding to one training cycle), we select $B$ samples from the unlabelled set $\mathcal{D}_U = \{ \xb_u ^{(j)} \}_{j = 1}^{M}$. These samples are annotated by an oracle before being added to the labelled training set $\mathcal{D}_L$. The new labelled and unlabelled sets \revisiontwo{are updated such that} $|\mathcal{D}_L| = N + B$ and $|\mathcal{D}_U| = M - B$. This iterative process is repeated until the total annotation budget is exhausted.

Our AL method addresses the problem of uncertainty-based strategies, generally prone to query samples with redundant information, in a simple and \revisiontwo{computationally}-efficient way. \revisiontwo{It builds upon our use of stochastic batches and operates in two stages to ensure a guided sampling diversity, summarized in Fig.~\ref{fig:stochastic_batches}}. First, we generate a pool of $Q$ batches, each containing $B$ samples chosen uniformly at random from $\mathcal{D}_U$:

\begin{equation}
    Batch^{(i)} = \{\xb_u^{(i_1)}, \xb_u^{(i_2)}, ..., \xb_u^{(i_B)}\} \sim Uniform(\mathcal{D}_u, B)
\end{equation}

For each generated batch, an uncertainty score is assigned to each unlabelled sample it contains, according to the current model $f_{\hat \theta}$ and the chosen uncertainty metric ($Uncert$):

\begin{equation}
\forall k = 1, ..., B:\quad 
 u_{score}^{x_u^{(i_k)}} = Uncert\ (f_{\hat \theta}, \xb_u^{(i_k)}).
 \end{equation}

The mean $u_{score}$ \revisiontwo{is computed across each generated} batch:
 
\begin{equation}
u_{score}^{Batch^{(i)}} = \frac{1}{B} \sum_{k = 1}^{B} u_{score}^{x_u^{(i_k)}}.
\end{equation}

The batch with the highest mean score yields the set of annotation candidates $X_{candidate}$, such that:

\begin{equation}
 X_{candidate} \leftarrow \argmax_{Batch^{(i)}} \left(u_{score}^{Batch^{(i)}}\right). 
\end{equation}


The algorithm for our stochastic batch selection strategy is presented in Alg.~\ref{alg:AL_stochasticbatches}.

\begin{algorithm}
\caption{\revisiontwo{Uncertainty-based sampling with Stochastic Batches}}
\label{alg:AL_stochasticbatches}
 \hspace*{\algorithmicindent} \textbf{Input} $\mathcal{D}_u,\ Q,\ B$
\begin{algorithmic}[1]
\For{$\xb_u \in \mathcal{D}_u$}
  \State $u_{score} \gets Uncert(f_{\hat \theta}, \xb_u)$\;
\EndFor
\For{$i \leftarrow 1$ to $Q$}
    \State$Batch^{(i)} = \{\xb_u^{(1)}, ..., \xb_u^{(B)}\} \gets Uniform(\mathcal{D}_u, B) $\;
    \State $u_{score}^{Batch^{(i)}} \gets$ Mean $u_{score}$ over all samples in $Batch^{(i)}$\;
\EndFor
\State $X_{candidate} \leftarrow \argmax_{Batch^{(i)}} \left(u_{score}^{Batch^{(i)}}\right)$
\end{algorithmic}
 \hspace*{\algorithmicindent} \textbf{Return}  $X_{candidate}$
\end{algorithm}

\section{Experiments}
\label{sec:experiments}

We assess the benefits of our proposed stochastic batches on a medical image segmentation task. Our evaluation compares the performance with and without stochastic batches of models trained with different uncertainty-based AL strategies. These strategies include Entropy-based sampling \citep{shannon_mathematical_1948}, Dropout-based sampling \citep{gal_dropout_2016}, Test-time augmentation (TTA)-based sampling \citep{gaillochet_TAAL_2022} and sampling based on Learning Loss \citep{yoo_learning_2019}, defined in more details in Sec.~\ref{par:sampling_baselines}. We start by evaluating the gains of our stochastic batch sampling on two medical image datasets.
We then assess the robustness of our method to the training and sampling procedure through a series of ablation studies on the initial labelled set size, training hyper-parameters, sampling budget and stochastic pool size.


\subsection{Datasets}

\revisiontwo{We validate our method on} two complementary datasets with different types of challenges: 1) the Prostate MR Image Segmentation (PROMISE) 2012 challenge \citep{litjens_evaluation_2014}, for prostate segmentation, with varying degrees of pixel intensity distributions \revisiontwo{(as pictured in Fig.~\ref{fig:example_ALsamples})}, and 2) the Medical Segmentation Decathlon \citep{antonelli_medical_2022} for the segmentation of anterior and posterior hippocampus, with varying degrees of anatomical shapes. 

The PROMISE12 dataset contains \revisiontwo{MRI} data from 50 patients, both healthy (or with benign diseases) and pathological (with prostate cancer).  
Each volume is converted to 2D images by slicing along the short axis. Images are then resampled to $1.0$\,mm isotropic resolution and resized to $128 \times 128$ pixels. 

Similarly, the Medical Segmentation Decathlon contains hippocampus data from 260 patients. The MRI volumes are converted to 2D images, which are resized to $50 \times 50$ pixels while kept to the original $1.0$\,mm isotropic resolution.  
The pixel intensity of both datasets \revisiontwo{is normalized based on the 1\% and 99\% percentiles for each scan}. 

We test our model on 10 patient volumes from the prostate dataset and 50 from the hippocampus dataset, all selected uniformly at random. \revisiontwo{This yields 248 and 1757 test images, respectively}. Our validation uses 109 prostate images composing 5 volumes, and 350 hippocampus images composing 10 volumes.  
Since active learning aims to minimize the amount of labelled data, we only use this validation set for hyper-parameter search purposes. Our ablation studies show that our method remains advantageous under different hyper-parameter settings. Our training set, labelled and unlabelled, comprises 1020 prostate images from 35 patients and 7163 hippocampus images from 200 patients.

\subsection{Evaluation metrics}
We evaluate our method on test volumes (3D) and individual images from these volumes (2D). We use both pixel overlap-based metrics and distance-based metrics. 

In terms of overlap-based metrics, we use the well-known Dice similarity coefficient (DSC), which ranges from 0\% (zero overlap) to 100\% (perfect overlap):

\begin{equation} 
    \mathrm{DSC}(X, Y) \, = \, \frac{2 | X \cap Y|}{|X| \cup |Y|}
\end{equation}

In our results, we report the DSC averaged over all non-background channels. 

The Hausdorff distance (HD) measures the quality of the segmentation by computing the maximum shortest distance between a point from the prediction contour and a point from the target contour. Since the Hausdorff distance tends to be sensitive to outliers, we use a more robust variant which considers the $95^{th}$ percentile instead of the true maximum (HD95).
Given $\mathrm{d}(x, Y)$ the minimum distance from the boundary pixel $x$ to the region $Y$, we get:

\begin{align} 
    \mathrm{HD95}(X, Y) & \, = \, \max \left\{ 95^{th}_{x \in X}\, \mathrm{d}(x, Y), \, 95^{th}_{y \in Y} \,\mathrm{d}(X, y) \right\} 
\end{align}

\subsection{Implementation details}
Medical annotations \revisiontwo{for image segmentation} are typically performed on \revisiontwo{all slices of a given image volume} \citep{ozdemir_active_2021}. However, to optimize the limited annotation resources, we conduct slice-based active learning and select individual images for annotation after every cycle. 
We start each experiment by training our model with 10 labelled images, randomly sampled from the unlabelled set before annotation. Setting the budget to $B = 10$, we use our AL strategy to select 10 new samples from the unlabelled set, annotate them and add them to the existing labelled set. This process corresponds to the first AL cycle, which we repeat for a fixed number of cycles. Similarly to the experimental setting of previous studies, we retrain the model from scratch after each AL cycle to evaluate model performance in a consistent way \citep{budd_survey_2021}.

Random processes such as model initialization or data shuffling are seeded. We repeat each experimental setup with 5 different seeds and report the mean and standard deviation of these runs as our result. Experiments were run on NVIDIA V100 and A6000 GPUs, with CUDA 10.2 and CUDA 12.0, respectively. We implement the methods using Python 3.8.10 with the PyTorch framework.

\subsubsection{Training}
State-of-the-art methods in medical image segmentation have often adopted UNet-based architectures \citep{ronneberger_u-net_2015}. Accordingly, we use a standard 4-layer UNet as a proxy for widely used architectures in our segmentation model, with dropout ($p=0.5$), batch normalization and a leaky ReLU activation function. \revisiontwo{Employing} such a model also focuses the evaluation on the improvement due to our stochastic batch strategy instead of measuring the performance of a backbone. However, without loss of generality, the use of alternative segmentation models could also be envisioned for our AL approach.

The model is trained for 75 epochs in all experiments, each iterating over 250 batches (training samples can appear in several batches), with a batch size of $4$. \revisiontwo{Training is hence carried out for a fixed $75 \times 250 = 18,750$ steps in} all experiments, ensuring a fairer comparison of model performance between AL cycles.

We optimize a supervised CE loss with the Adam optimizer \citep{kingma_adam_2015}. We apply a gradual warmup with a cosine annealing scheduler \citep{loshchilov_sgdr_2017,goyal_accurate_2018} to control the learning rate. During training, we use data augmentations on the input, with parameters $d$ and $\epsilon$, where $d$ is the degree of rotation in 2D, and $\epsilon$ models Gaussian noise.

When not testing for their impact, we keep the training hyper-parameters fixed. We fix the initial learning rate $LR = 10^{-6}$ with optimizer weight decay set to $10^{-4}$. The scheduler increments the learning rate by a factor $200$ during the first $10$ epochs.  
For augmentations, we set $d \sim \mathcal{U}(-10, 10)$ and $\epsilon \sim \mathcal{N}(0, 0.01)$.

Since active learning aims to minimize the amount of labelled data needed to train the model, we minimize the use of the validation set and avoid its use to select the final model. Our final model is instead the model obtained after the last training epoch.

\subsubsection{Active learning sampling}

\paragraph{Baselines}
\label{par:sampling_baselines}
We compare our stochastic batches strategy with random sampling (RS), \revisiontwo{Core-set} \citep{sener_active_2018}, and four purely uncertainty-based methods: 

\begin{itemize}
    \item Entropy-based uncertainty \citep{shannon_mathematical_1948}, which computes the entropy on the predicted output probabilities:
    {\small $$Uncert\ (f_{\hat \theta} , \xb_u^{(i_k)}) = - \sum_i p(y_i | \xb_u^{(i_k)}, \hat \theta)\ \log p(y_i | \xb_u^{(i_k)}, \hat \theta);$$}
    
    \item Dropout-based uncertainty \citep{gal_dropout_2016}, using the divergence of $K$ predictions obtained by multiple inferences with dropout $d$:
    {\small $$Uncert\ (f_{\hat \theta} , \xb_u^{(i_k)}) = Div \big(f_{\hat \theta, d_1}(\xb_u^{(i_k)}),\ ...,\ f_{\hat \theta, d_K}(\xb_u^{(i_k)}) \big);$$}
    
    \item Test-time Augmentation (TTA)-based uncertainty \citep{gaillochet_TAAL_2022}, which measures the divergence of predictions obtained for $K$ transformations $\Gamma$ to the input:
    {\small $$Uncert\ (f_{\hat \theta} , \xb_u^{(i_k)}) = Div\ \big(\Gamma_1^{-1}[f_{\hat \theta}(\Gamma_1(\xb_u^{(i_k)}))],\ ...,\ \Gamma_{K}^{-1}[f_{\hat \theta}(\Gamma_{K}(\xb_u^{(i_k)}))]\big);$$}
    
    \item Learning Loss (LL) uncertainty \citep{yoo_learning_2019}, which trains an external module $L_{\tilde \theta}$ to predict the target losses from a feature set $h$ extracted from the hidden layers of $f_{\hat \theta}$:
    {\small $$Uncert\ (f_{\hat \theta} , L_{\tilde \theta}, \xb_u^{(i_k)}) = L_{\tilde \theta}(h(\xb_u^{(i_k)})).$$}
\end{itemize}

These purely uncertainty-based methods query batches made of the most uncertain samples according to a sample-level uncertainty measure.

Similarly to \cite{gaillochet_TAAL_2022}, as our divergence measure for Dropout-based and TTA-based sampling, we use a standard Jensen–Shannon divergence (JSD) on the output probability maps obtained from $K=8$ inferences. For TTA, augmentations $\Gamma$ include Gaussian noise $\epsilon \sim \mathcal{N}(0, 0.01)$ and rotation. To simulate more realistic transformations in medical data, we replace the 90 degrees rotations in \cite{gaillochet_TAAL_2022} with rotations of angle $ d \sim  \mathcal{U}(-10, 10)$ degrees.
The training parameters used for the approach based on Learning Loss \citep{yoo_learning_2019} were obtained by grid search on 10 labelled samples. We kept these parameters fixed in all our experiments. 

\paragraph{Stochastic batches}
We generate the pool of stochastic batches by iteratively sampling $B$ unlabelled images uniformly at random and without replacement. In other words, we divide the unlabelled samples into $Q$ pools of $B$ samples. Hence the stochastic pool has size $Q = \mathrm{floor}\big(\nicefrac{|\mathcal{D}_U|}{B})$, and it reduces in size with the number of AL cycles.

\section{Results}

\subsection{AL performance on the Prostate and Hippocampus \revisiontwo{datasets}}
\label{subsec:results_overall}

\begin{table*}[htb]
\setlength{\tabcolsep}{3.2pt} 
\caption{\textbf{Overall improvements with Stochastic Batches over varying initial labelled samples}. Mean model performance over all AL cycles. We show the mean (std) Dice score (DSC, higher is better) and $95\%$ Hausdorff distance (HD95, lower is better) over 3D test volumes and 2D test images. The results are averaged over 5 initial labelled sets chosen uniformly at random and 6 AL cycles (we omit results with the initial labelled set as they are similar across all methods). A * indicates the statistical significance of the result with a p-value $< 0.05$ given a paired permutation test.
}
\centering
    \small 
        \begin{tabular}{cc  ccc ccc ccc}
        \toprule
          & & \multicolumn{3}{c}{\textbf{Prostate}} &  \multicolumn{3}{c}{\textbf{Anterior Hippocampus}} &  \multicolumn{3}{c}{\textbf{Posterior Hippocampus}} \\
          \cmidrule(l{5pt}r{5pt}){3-5}
          \cmidrule(l{5pt}r{5pt}){6-8}
          \cmidrule(l{5pt}r{5pt}){9-11}
        & & \textbf{3D DSC} & \textbf{2D DSC} & \textbf{3D HD95} & \textbf{3D DSC} & \textbf{2D DSC} & \textbf{3D HD95} & \textbf{3D DSC} & \textbf{2D DSC} & \textbf{3D HD95}\\
        \midrule
        \multirow{2}{*}{\textbf{RS}} & & 68.83 & 67.94  & 7.032 &
                                        77.42 & 75.45 & 4.09 & 
                                        76.43 & \textbf{70.02} & 4.51 \\
         &  & \ppm{15.99} & \ppm{8.28} & \ppm{3.734} & 
                \ppm{1.67} & \ppm{1.13} & \ppm{0.47} & 
                \ppm{0.80} & \ppm{1.62} & \ppm{0.70}\\
         \midrule 
        \textbf{Core-set}
        &  & 68.84 & 65.87 & 7.64 &
                                            78.83 & 73.14 & 4.45 & 
                                            75.32 & 66.45 & 4.52\\
         \scriptsize\citep{sener_active_2018} &  & \ppm{17.37} & \ppm{7.31} & \ppm{2.73} & 
                \ppm{3.25} & \ppm{1.20} & \ppm{0.46} & 
                \ppm{5.46} & \ppm{1.29} & \ppm{0.53}\\
         \midrule 
        \multirow[b]{2}{*}{\textbf{Entropy}} & \multirow[t]{2}{*}{\footnotesize w/o SB} & 67.01 & 66.88 & 7.026 & 
                            78.22 & 75.03 & 3.79 & 
                            74.68 & 65.70 & 5.10\\
        & & \ppm{16.68} & \ppm{8.62}  & \ppm{4.271} &   
                \ppm{1.90} & \ppm{0.97} & \ppm{0.23} & 
                \ppm{1.60} & \ppm{1.66} & \ppm{1.10}\\
        \cdashline{2-11}
        \rule{0pt}{3ex}  
        \multirow[t]{2}{*}{\scriptsize\citep{shannon_mathematical_1948}} & \multirow[t]{2}{*}{\footnotesize Ours} & 71.27* & 68.99* & 6.689* & 
                            79.25* & 75.84 & \textbf{3.72} & 
                            76.23* & 69.01* & \textbf{3.85}*\\
         & & \ppm{17.39} & \ppm{9.03} & \ppm{3.143}& 
                \ppm{0.86} & \ppm{0.86} & \ppm{0.15} & 
                \ppm{0.87} & \ppm{1.97} & \ppm{0.31}\\
         \midrule 
        \multirow[b]{2}{*}{\textbf{Dropout}} & \multirow{2}{*}{\footnotesize w/o SB} & 67.69 & 67.07 & 6.964 & 
                            78.22 & 74.29 & 4.04 & 
                            74.45 & 66.78 & 4.77\\
        & & \ppm{17.16} & \ppm{9.51} & \ppm{4.952} & 
                \ppm{1.28} & \ppm{1.10} & \ppm{0.33} & 
                \ppm{1.20} & \ppm{2.06} & \ppm{1.19}\\
        \cdashline{2-11}
        \rule{0pt}{3ex}  
        \multirow[t]{2}{*}{\scriptsize\citep{gal_dropout_2016}} & \multirow{2}{*}{\footnotesize Ours} & \textbf{72.59*} & \textbf{69.64*} & \textbf{6.583*} & 
                            \textbf{79.28*} & \textbf{76.36*} & 3.73 & 
                            76.27* & \textbf{68.94*} & \textbf{3.88*}\\
         & & \ppm{14.96} & \ppm{8.05} & \ppm{3.177} & 
            \ppm{0.83} & \ppm{0.69} & \ppm{0.10} & 
            \ppm{0.85} & \ppm{1.15} & \ppm{0.39}\\
         \midrule 
        \multirow[b]{2}{*}{\textbf{TTA}} & \multirow{2}{*}{\footnotesize w/o SB} & 64.07 & 65.85 & 6.918* & 
                            77.31 & 73.66 & 4.10 & 
                            73.84 & 64.94 & 5.07\\
        & & \ppm{21.13} & \ppm{10.25} & \ppm{4.794} & 
            \ppm{3.24} & \ppm{1.08} & \ppm{0.54} & 
            \ppm{2.01} & \ppm{0.59} & \ppm{0.93}\\
        \cdashline{2-11}
        \rule{0pt}{3ex}  
        \multirow[t]{2}{*}{\scriptsize\citep{gaillochet_TAAL_2022}}  & \multirow{2}{*}{\footnotesize Ours} & 69.71* & 68.00* & 7.188 &
                            78.86* & 75.25 & 4.07 & 
                            \textbf{76.44*} & 67.08* & 4.43*\\
         & & \ppm{17.59} & \ppm{9.02} & \ppm{3.173} & 
            \ppm{0.94} & \ppm{1.41} & \ppm{0.31} & 
            \ppm{0.90} & \ppm{1.01} & \ppm{0.40}\\
        \midrule
        \multirow[b]{2}{*}{\textbf{Learning Loss}} & \multirow{2}{*}{\footnotesize w/o SB} & 53.88 & 60.22 & 9.139 & 
                            62.54 & 69.70 & 5.94 & 
                            61.57 & 62.82 & 5.87\\
        & & \ppm{21.51} & \ppm{10.36} & \ppm{6.439} &
            \ppm{1.38} & \ppm{0.92} & \ppm{0.59} & 
            \ppm{2.83} & \ppm{1.14} & \ppm{0.12} \\
        \cdashline{2-11}
        \rule{0pt}{3ex}  \multirow[t]{2}{*}{\scriptsize\citep{yoo_learning_2019}} & \multirow{2}{*}{\footnotesize Ours} & 65.29* & 65.72* & 7.816* & 
                            72.09* & 74.32* & 4.51* & 
                            71.23* & 67.75* & 5.16\\
         & & \ppm{17.72} & \ppm{8.94} & \ppm{4.384} & 
            \ppm{2.73} & \ppm{0.85} & \ppm{0.60} & 
            \ppm{1.29} & \ppm{1.21} & \ppm{0.63}\\
            \bottomrule
        \end{tabular}
\label{table:results_overall}
\end{table*}

We validate our proposed stochastic batch sampling strategy by looking at the AL performance over 5 different initial labelled sets chosen uniformly at random from the training set. 
Tab.~\ref{table:results_overall} shows the average results over all AL cycles for both Prostate and Hippocampus data. Note that the standard deviations given in the table tend to be large because they are averaged over multiple initial labelled sets, initialization seeds and AL cycles. For all methods and metrics except for TTA on Prostate with the $95\%$ Hausdorff distance metric, stochastic batch sampling constantly provides improved performance over its purely uncertainty-based counterpart, both in terms of overlap-based and distance-based metric. 

\begin{figure*}[htb]
    \centering
    \begin{subfigure}{0.5\textwidth}
        \centering
        \includegraphics[width=\linewidth]{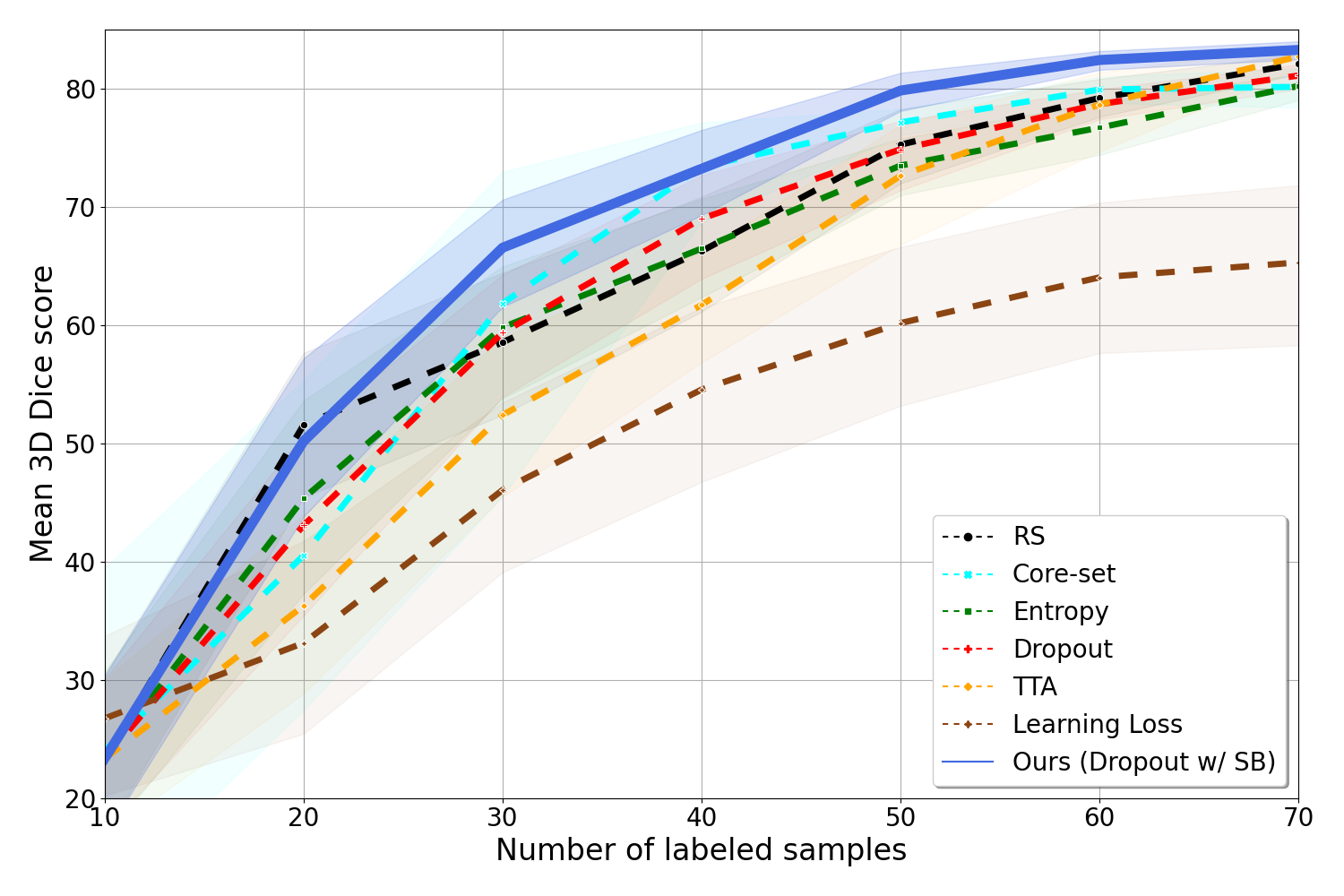}
        \caption{Prostate}
        \label{subfig:results_prostate_all}
    \end{subfigure}\hfil 
    \hfil
    \centering
    \begin{subfigure}{0.5\textwidth}
        \includegraphics[width=\linewidth]{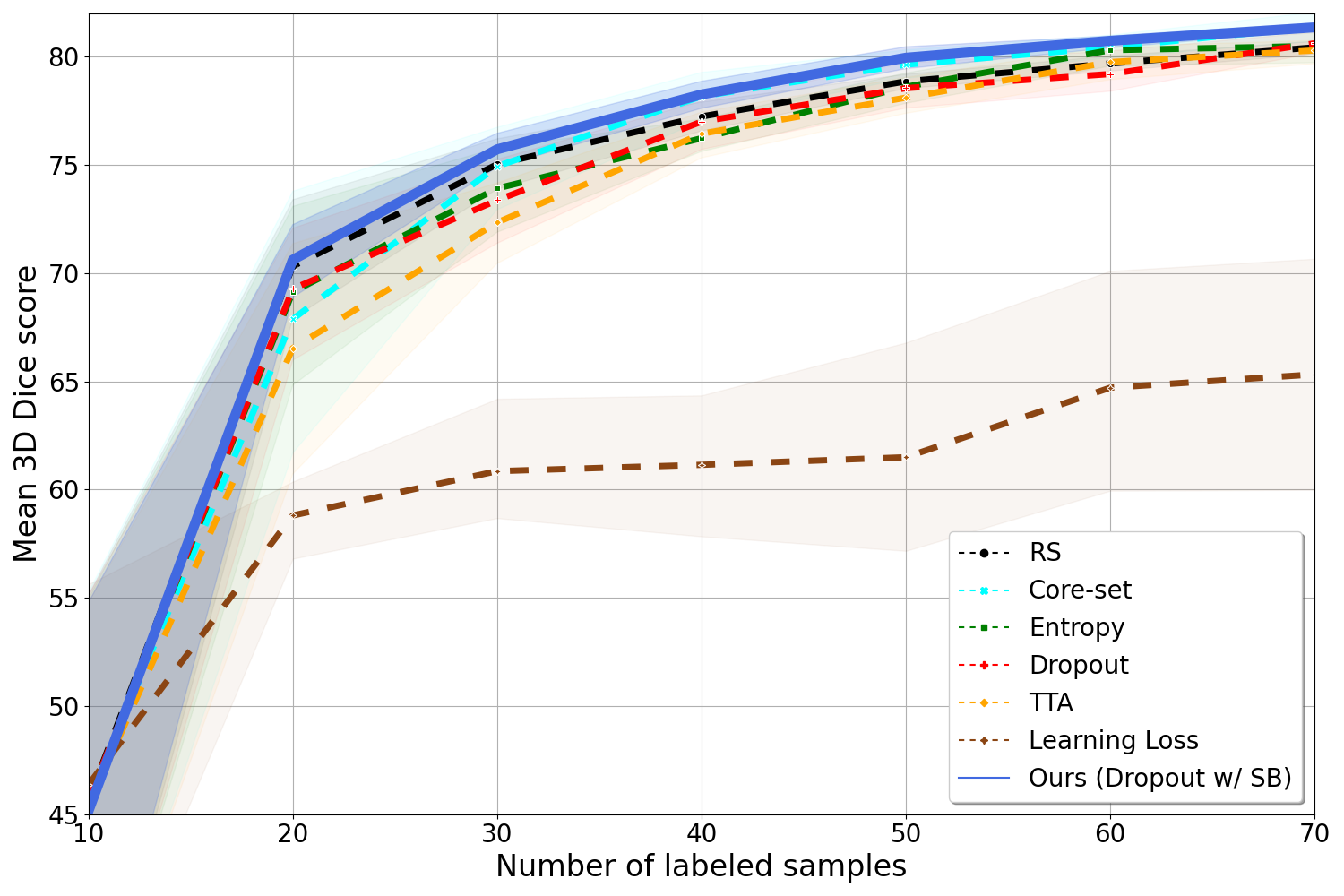}
        \caption{Hippocampus}
        \label{subfig:results_hippocampus_all}
    \end{subfigure}\hfil 
    \caption{\textbf{Overall AL performance on the \subref{subfig:results_prostate_all}) Prostate and \subref{subfig:results_hippocampus_all}) Hippocampus \revisiontwo{datasets}}. Our best stochastic batch sampling method (full-blue) outperforms all other methods, including \revisiontwo{Core-set} and random sampling (RS).}
    \label{fig:results_initlabeled_all}
\end{figure*}

We also observe that stochastic batch sampling outperforms both random sampling and \revisiontwo{Core-set} \citep{sener_active_2018}, a diversity-based AL approach. This is corroborated by Fig.~\ref{fig:results_initlabeled_all}, which show that Dropout with our stochastic batches outperforms all other baseline methods in terms of 3D dice score, in almost all AL cycles. In addition, Tab.~\ref{table:results_time} gives the average time required by each strategy to provide a candidate set for annotation from the Hippocampus dataset. We see that using stochastic batches does not increase the sampling time of uncertainty-based methods. Furthermore, sampling with our proposed method is always much faster than with \revisiontwo{Core-set}.

\begin{figure*}[htb]
    \centering
    \begin{subfigure}{0.25\textwidth}
        \includegraphics[width=\linewidth]{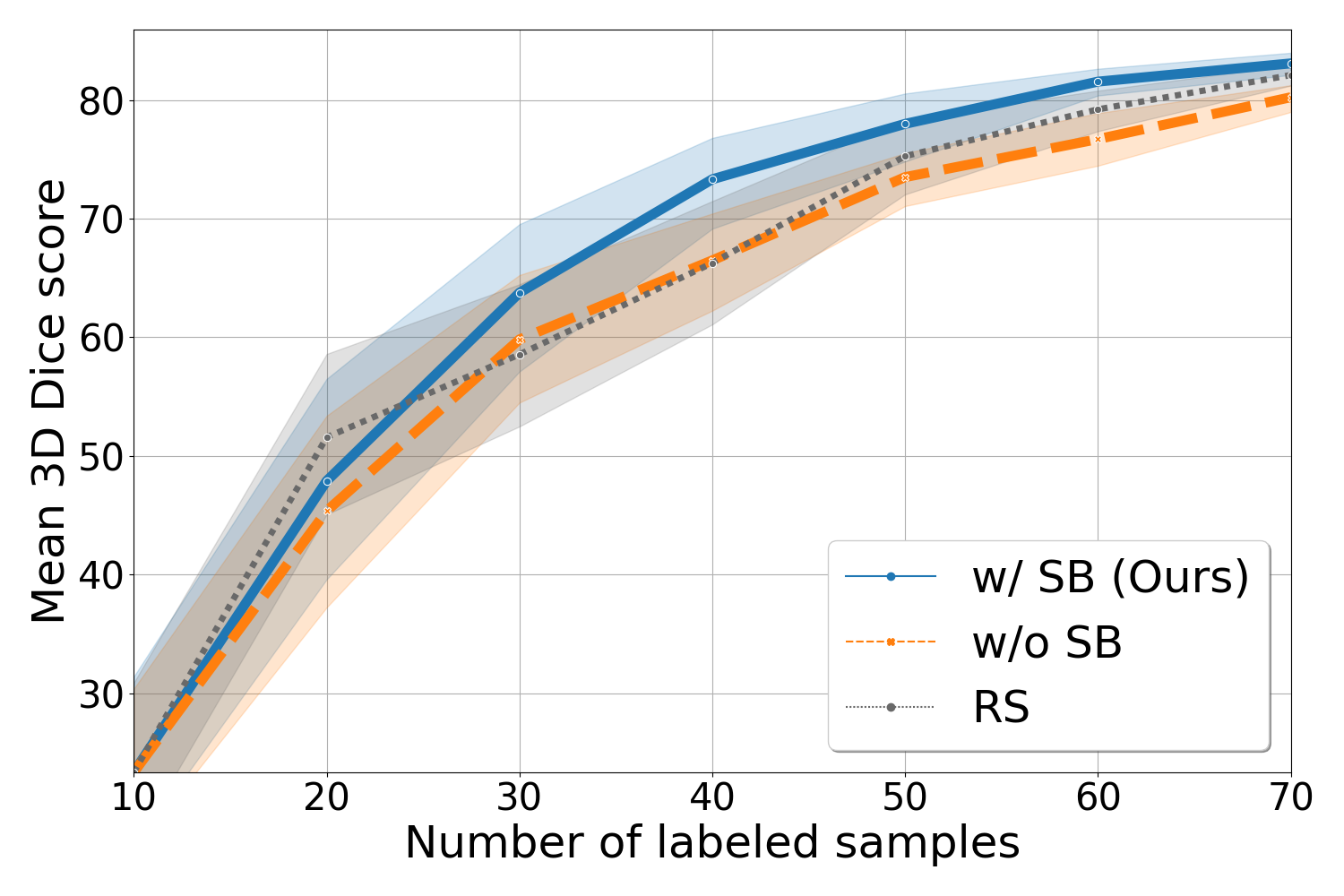}
        \caption{\small{Improvements for Entropy \\\citep{shannon_mathematical_1948} \label{fig:sub:results_initlabelled_entropy}}}
    \end{subfigure}\hfil 
    \hfil
    \centering
    \begin{subfigure}{0.25\textwidth}
        \includegraphics[width=\linewidth]{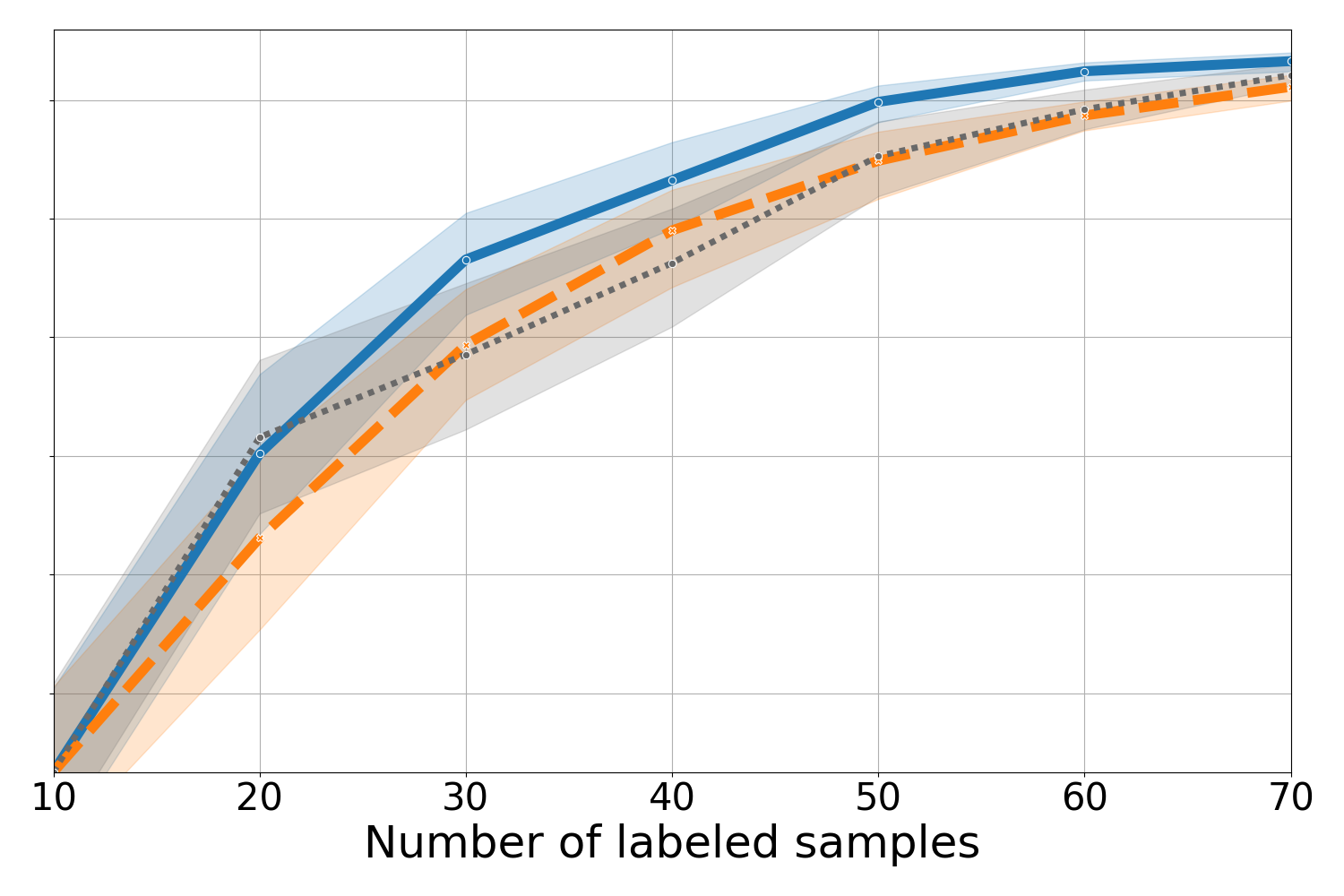}
        \caption{\small{Improvements for Dropout \\\citep{gal_dropout_2016} \label{fig:sub:results_initlabelled_dropout}}}
    \end{subfigure}\hfil 
    \medskip 
    \centering
    \begin{subfigure}{0.25\textwidth}
        \includegraphics[width=\linewidth]{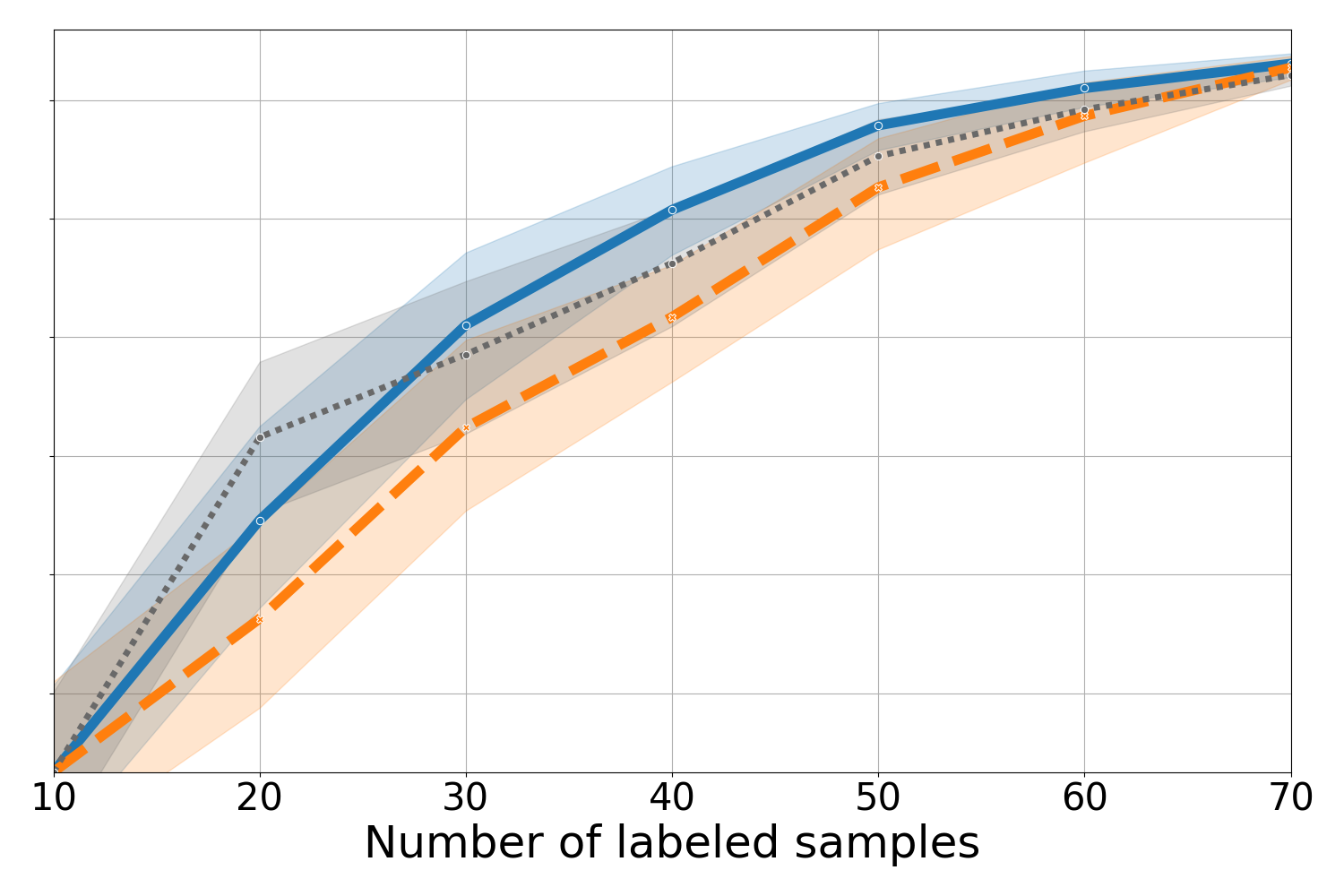}
        \caption{\small{Improvements for TTA \\\citep{gaillochet_TAAL_2022} \label{fig:sub:results_initlabelled_TTA}}}
    \end{subfigure}\hfil 
    \hfil
    \centering
    \begin{subfigure}{0.25\textwidth}
        \includegraphics[width=\linewidth]{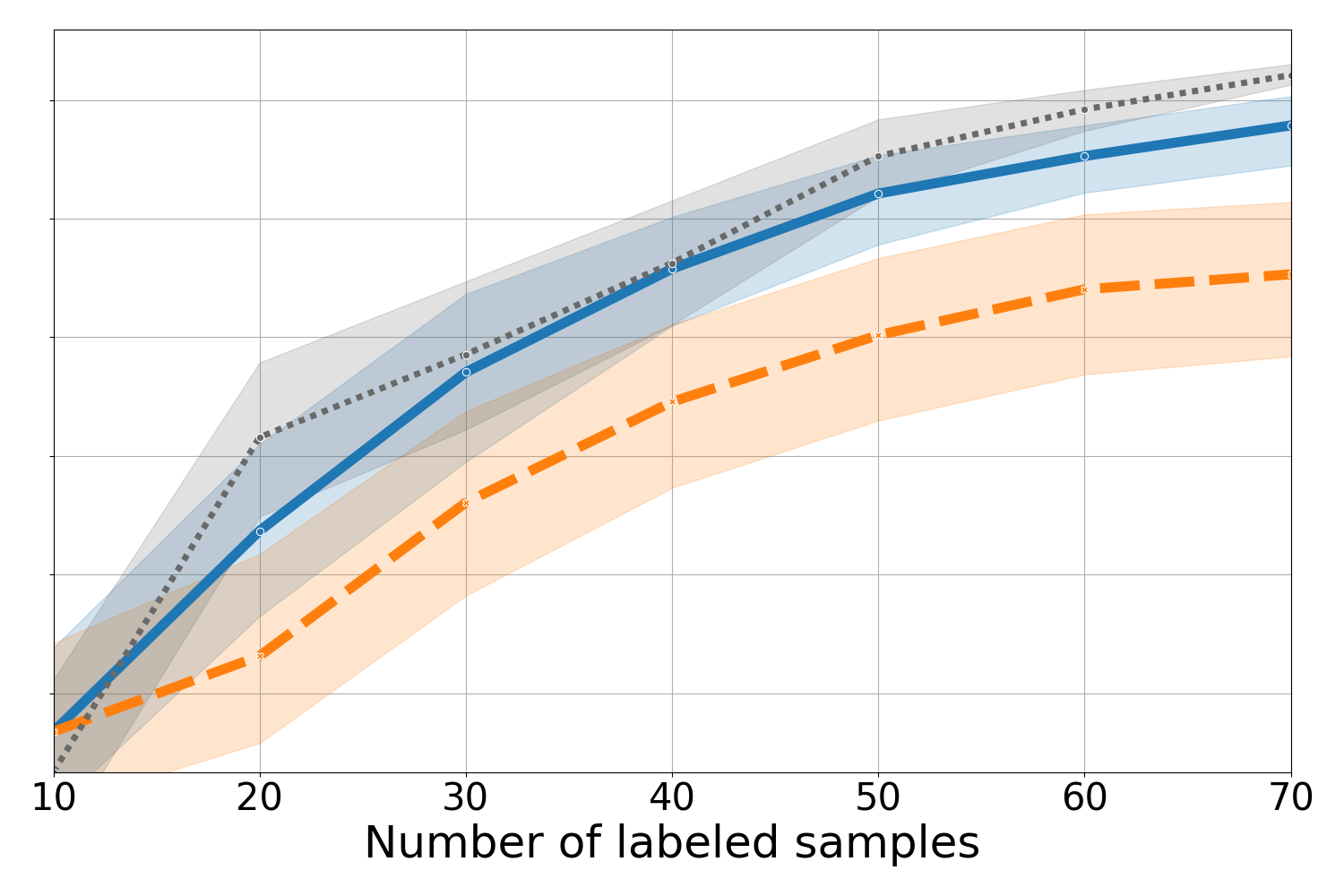}
        \caption{\small{Improvements for LL \\\citep{yoo_learning_2019} \label{fig:sub:results_initlabelled_learningloss}}}
    \end{subfigure}\hfil 
    \caption{\textbf{Individual improvements with Stochastic Batches on the Prostate \revisiontwo{dataset}}. Active learning results in terms of 3D test dice score and corresponding 95\% confidence interval. The results are averaged over 5 different initial labelled sets and 5 initialization seeds. Depicted are the results for sampling based on \subref{fig:sub:results_initlabelled_entropy}) Entropy, \subref{fig:sub:results_initlabelled_dropout}) Dropout, \subref{fig:sub:results_initlabelled_TTA}) Test-time augmentation and \subref{fig:sub:results_initlabelled_learningloss}) Learning Loss). The active learning selection is shown with (blue, full) and without (orange, dashed) stochastic batches, and random sampling is plotted in dotted grey. Stochastic batches improve the model performance of purely uncertainty-based AL strategies, regardless of the initial labelled set, repeatedly outperforming random sampling.
    }
    \label{fig:results_initlabelled}
\end{figure*}

\begin{figure*}[htb]
    \centering
    \begin{subfigure}{0.25\textwidth}
        \includegraphics[width=\linewidth]{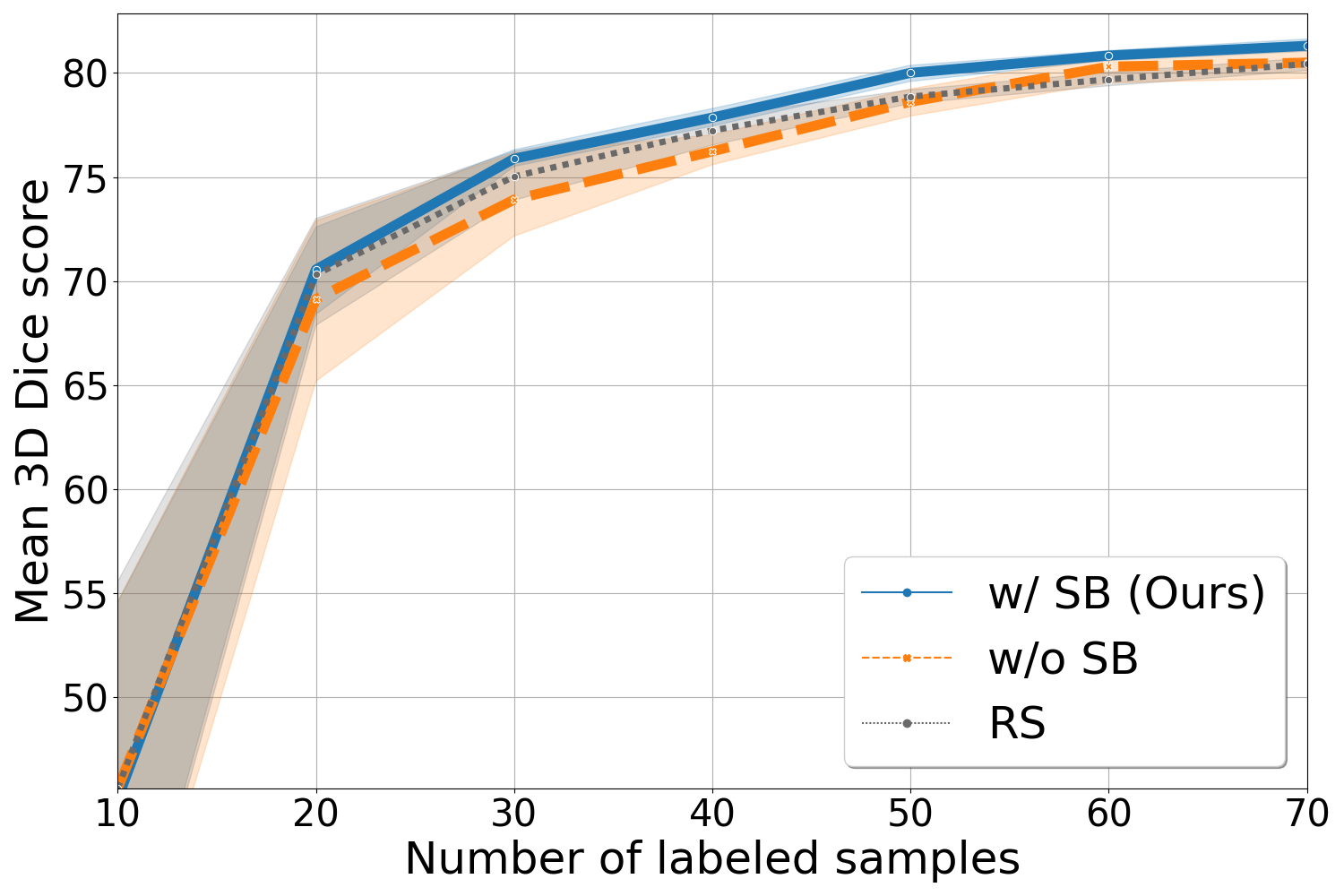}
        \caption{Improvements for Entropy \\\citep{shannon_mathematical_1948} \label{fig:sub:results_hippocampus_entropy}}
    \end{subfigure}\hfil 
    \hfil
    \centering
    \begin{subfigure}{0.25\textwidth}
        \includegraphics[width=\linewidth]{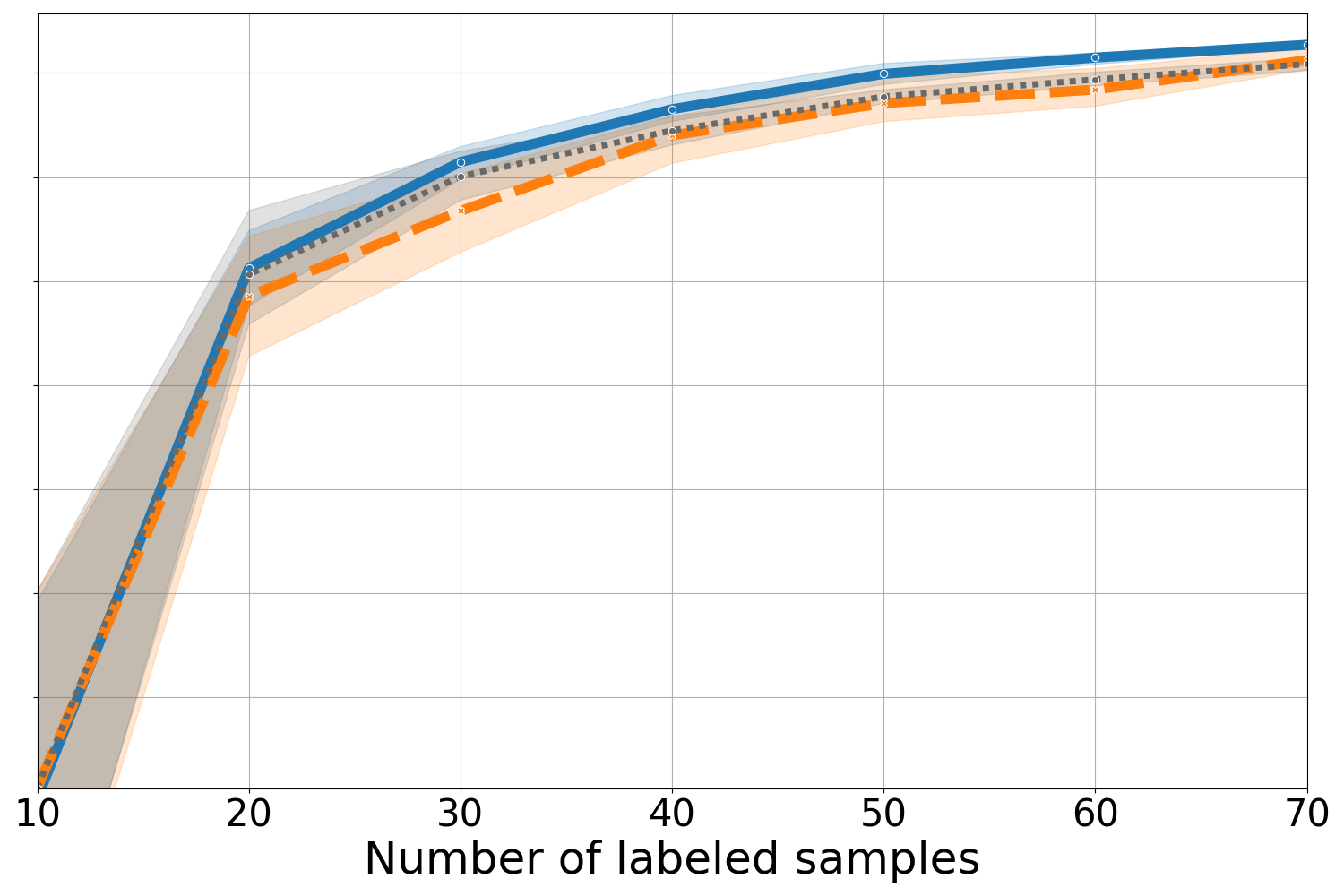}
        \caption{Improvements for Dropout \\\citep{gal_dropout_2016} \label{fig:sub:results_hippocampus_dropout}}
    \end{subfigure}\hfil 
    \medskip 
    \centering
    \begin{subfigure}{0.25\textwidth}
        \includegraphics[width=\linewidth]{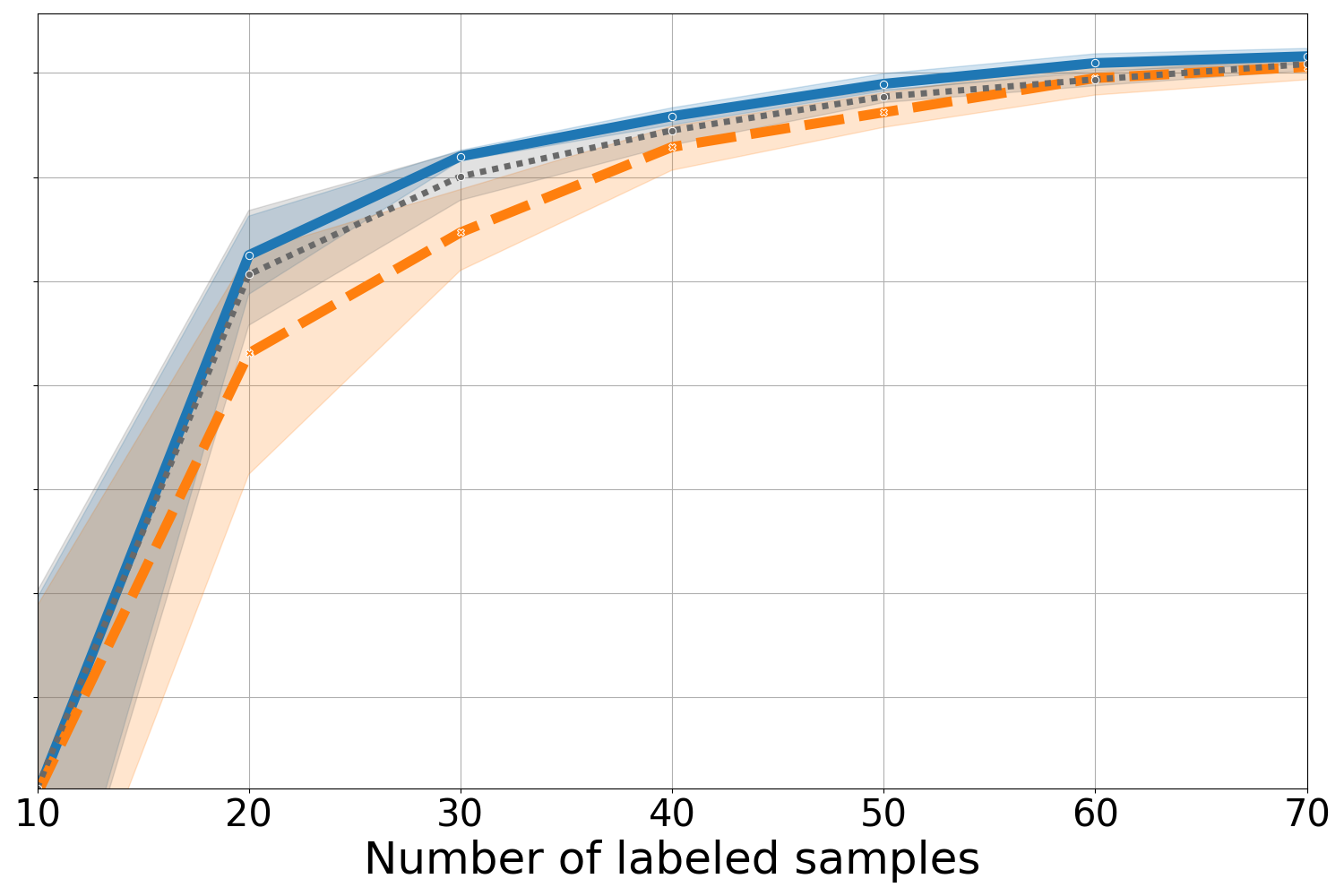}
        \caption{Improvements for TTA \\\citep{gaillochet_TAAL_2022} \label{fig:sub:results_hippocampus_TTA}}
    \end{subfigure}\hfil 
    \hfil
    \centering
    \begin{subfigure}{0.25\textwidth}
        \includegraphics[width=\linewidth]{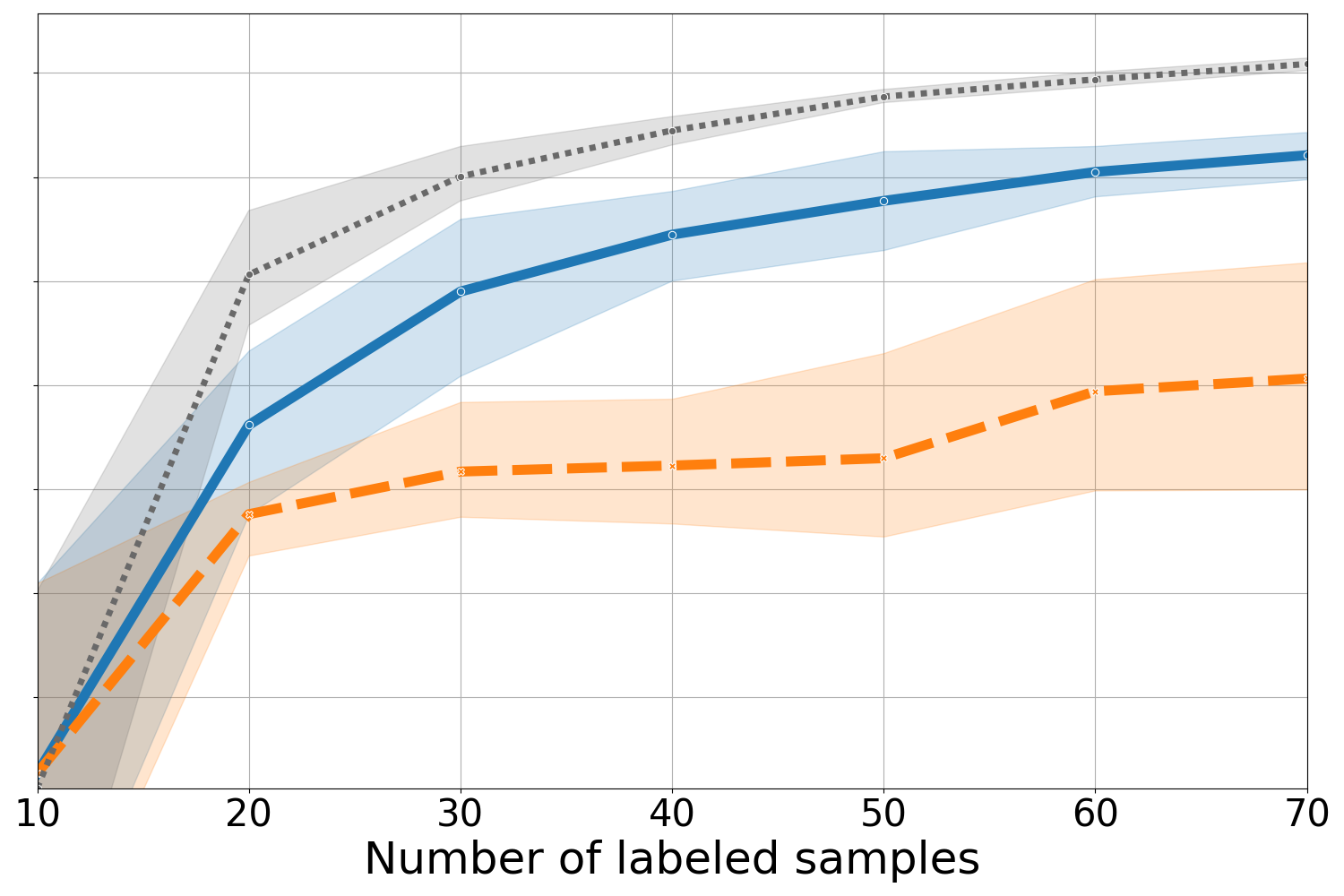}
        \caption{Improvements for LL \\\citep{yoo_learning_2019} \label{fig:sub:results_hippocampus_learningloss}}
    \end{subfigure}\hfil 
    \caption{\textbf{Individual improvements with Stochastic Batches on the Hippocampus \revisiontwo{dataset}}. Active learning results on the Hippocampus dataset in terms of 3D test dice score and corresponding 95\% confidence interval. The results are averaged over 5 different initial labelled sets. Depicted are the results for sampling based on \subref{fig:sub:results_hippocampus_entropy}) Entropy, \subref{fig:sub:results_hippocampus_dropout}) Dropout, \subref{fig:sub:results_hippocampus_TTA}) Test-time augmentation and \subref{fig:sub:results_hippocampus_learningloss}) Learning Loss. Sampling with Stochastic batches (blue, full) improves the model performance of purely uncertainty-based AL strategies (orange, dashed), regardless of the initial labelled set, boosting it above random sampling (grey, dotted) in the majority of cases.}
    \label{fig:results_hippocampus}
\end{figure*}

\begin{table*}[htb]
\caption{\textbf{Sampling time.} Mean sampling time computed over all AL cycles, for the Hippocampus dataset.}
{\small 
\begin{center}
\begin{tabular}{c c c cc cc cc cc}
\toprule
 & \multirow[c]{2}{*}{RS} & \multirow[c]{2}{*}{Core-set} & \multicolumn{2}{c}{Entropy} & \multicolumn{2}{c}{Dropout} & \multicolumn{2}{c}{TTA} & \multicolumn{2}{c}{Learning Loss} \\
 \cmidrule(l{5pt}r{5pt}){4-5}
 \cmidrule(l{5pt}r{5pt}){6-7}
 \cmidrule(l{5pt}r{5pt}){8-9}
 \cmidrule(l{5pt}r{5pt}){10-11}
&  &  & w/o \textsc{SB} & Ours & w/o \textsc{SB} & Ours & w/o \textsc{SB} & Ours & w/o \textsc{SB} & Ours\\
\midrule
 \multirow[c]{1}{*}{Time (min.)} & 0.00 & 0.71 & 0.12 & 0.11 & 0.58 & 0.58 & 0.37 & 0.37 & 0.16 & 0.18\\
 \bottomrule
\end{tabular}
\end{center}
}
\label{table:results_time}
\end{table*}

When looking at each dataset in more detail, the pairwise results on the Prostate dataset, shown in Fig.~\ref{fig:results_initlabelled}, validate the effectiveness of our method against different initial labelled sets. Averaged over 25 experiments with varying initial labelled sets and initialization seeds, our stochastic batch querying (blue, full lines) improves the model's performance of purely uncertainty-based strategies (orange, dashed lines). For all considered AL strategies, selecting the most uncertain batch of samples rather than the most uncertain individual samples improves the model's overall performance. The 3D dice score is always boosted, either over the score obtained by random sampling (grey, dotted) or to a level similar to that of a random sampling if the score were originally much lower, such as in the case of the Learning Loss. Indeed, Learning Loss has noticeably lower performance compared to Entropy, Dropout and TTA-based sampling. The Learning Loss approach involves backpropagating the gradient through both the task model and loss module during training. Both are updated \revisiontwo{simultaneously}, which means that training the loss module affects the training of the task model and vice versa. For comparability reasons and following most works in AL, we tuned the hyper-parameters such that the best validation performance was obtained on the first AL cycle (with the initial labelled set). We believe this could explain the poorer performance of Learning Loss with an increasing number of labelled samples. Similar observations can be made from Hippocampus data, as shown in Fig.~\ref{fig:results_hippocampus}.

\begin{figure}[h!]
    \centering
    \setlength{\tabcolsep}{2pt}
    \begin{small}
    \begin{tabular}{ccccc}
    \figBatches{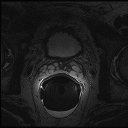} &   
    \figBatches{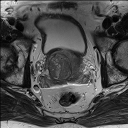} & 
    \phantom{-} &
    \figBatches{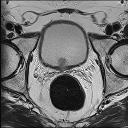} &
    \figBatches{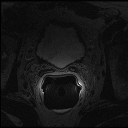} \\
    \figBatches{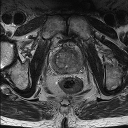} & 
    \figBatches{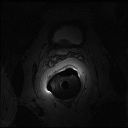} & 
    \phantom{-} &
    \figBatches{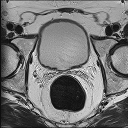} &
    \figBatches{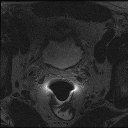} \\
    \figBatches{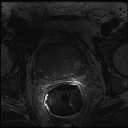} & 
    \figBatches{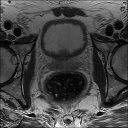} & 
    \phantom{-} &
    \figBatches{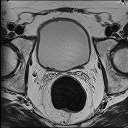} &
   \figBatches{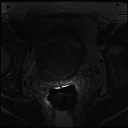} \\
   \figBatches{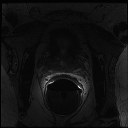} & 
   \figBatches{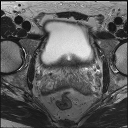} & 
   \phantom{-} &
    \figBatches{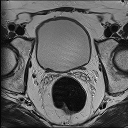} & 
    \figBatches{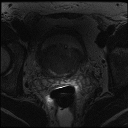} \\
    \figBatches{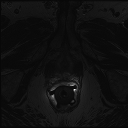} & 
    \figBatches{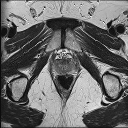} & 
    \phantom{-} &
    \figBatches{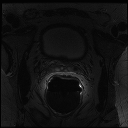} & 
   \figBatches{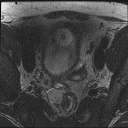} \\
   \multicolumn{2}{c}{(a) Ours} & 
    \phantom{-} &
   \multicolumn{2}{c}{(b) w/o Stochastic Batches}
    \end{tabular}
    \end{small}
    \caption{\textbf{Candidate batches from the \revisiontwo{Prostate dataset}}. The samples were selected by Entropy-based AL sampling with (first two columns) and without (last two columns) stochastic batches. While the candidate batch obtained via purely uncertainty-based sampling contains similar samples, selection with stochastic batches reduces the number of redundancies.}
    \label{fig:example_ALsamples}
\end{figure}

\begin{figure}[h!]
    \centering
    \setlength{\tabcolsep}{3pt}
    \begin{footnotesize}
    \begin{tabular}{ccccc}
    \multirow{3}{*}[35pt]{\figSeg{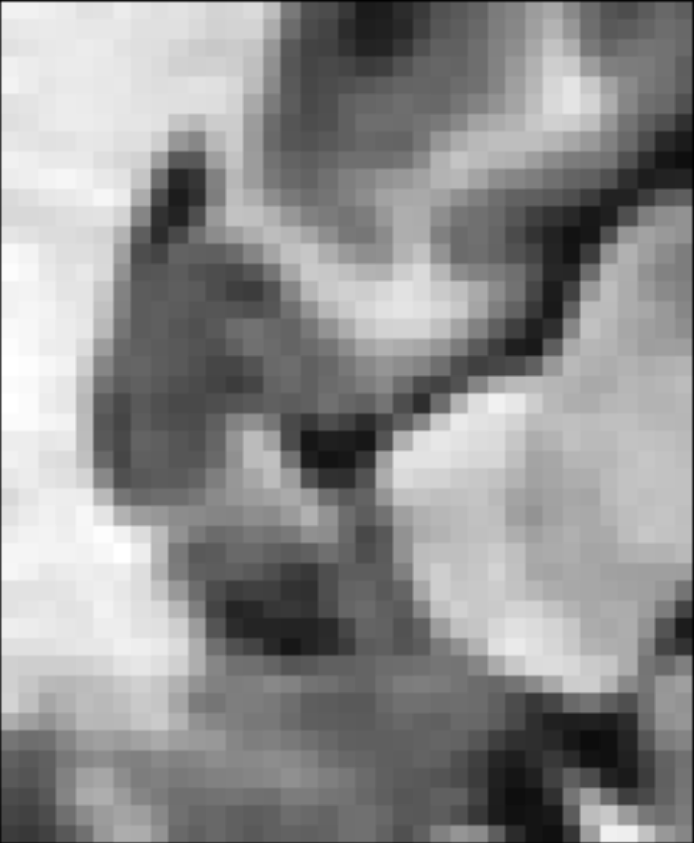}} &
    \multirow{1}{*}[25pt]{\revisiontwo{\shortstack[r]{Ours:\\ (PH\,/\,AH)}}} &
    \figSeg{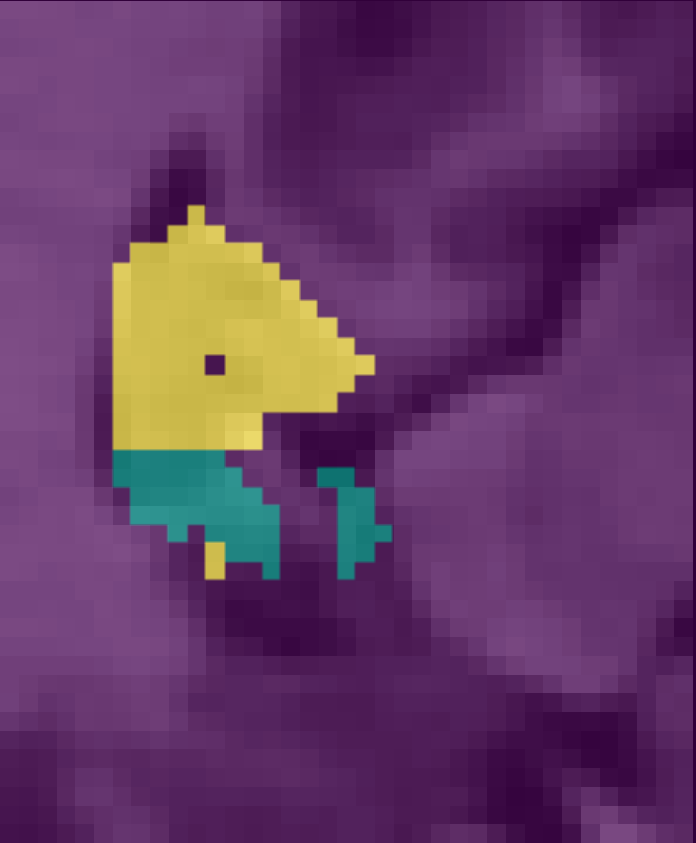} &
    \figSeg{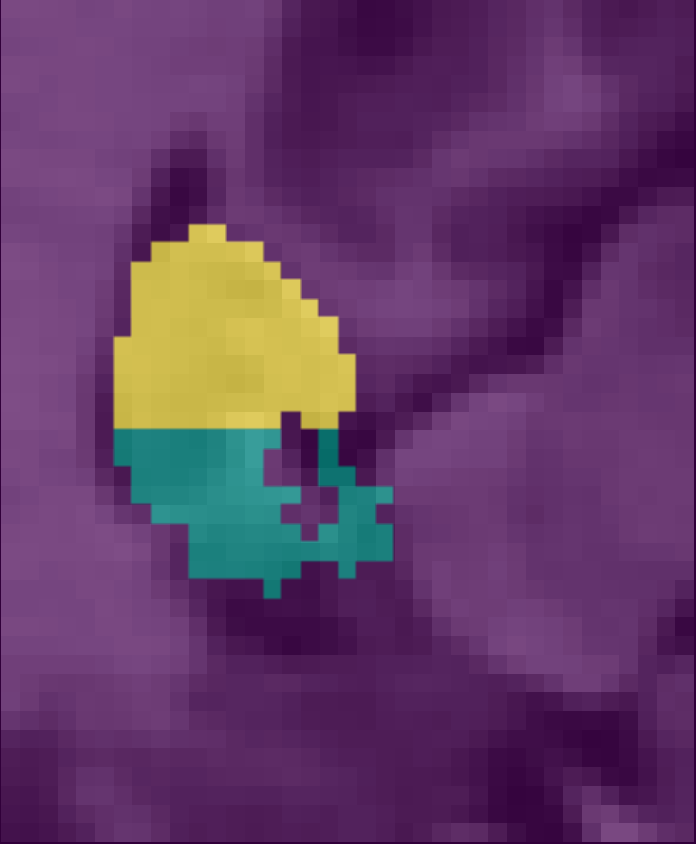} &
    \figSeg{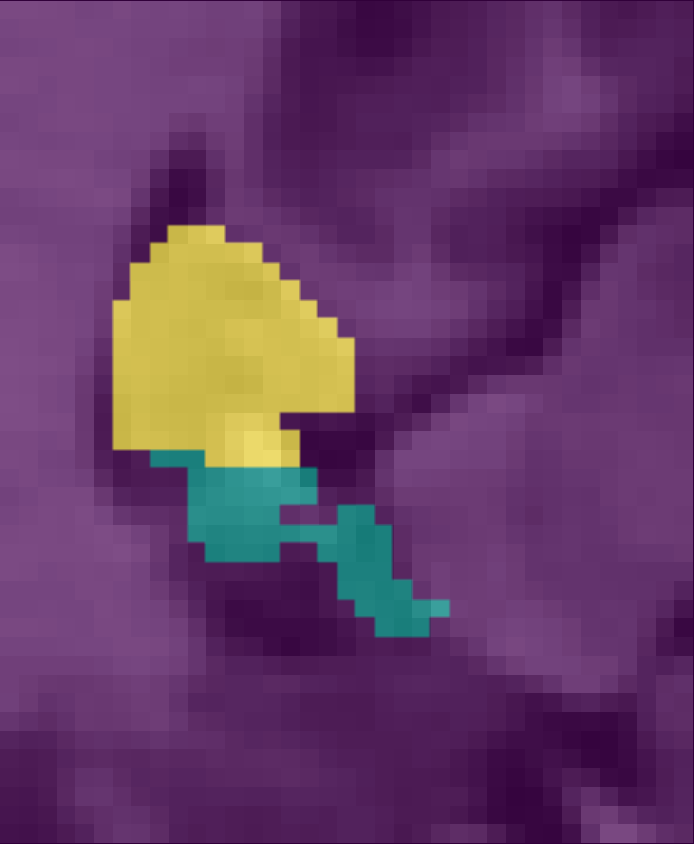} \\
     &  & \revisiontwo{89.26\,/\,56.20} & \revisiontwo{87.72\,/\,65.33} & \revisiontwo{93.06\,/\,69.23}\\[3pt]
    
    \multirow{3}{*}[50pt]{\figSeg{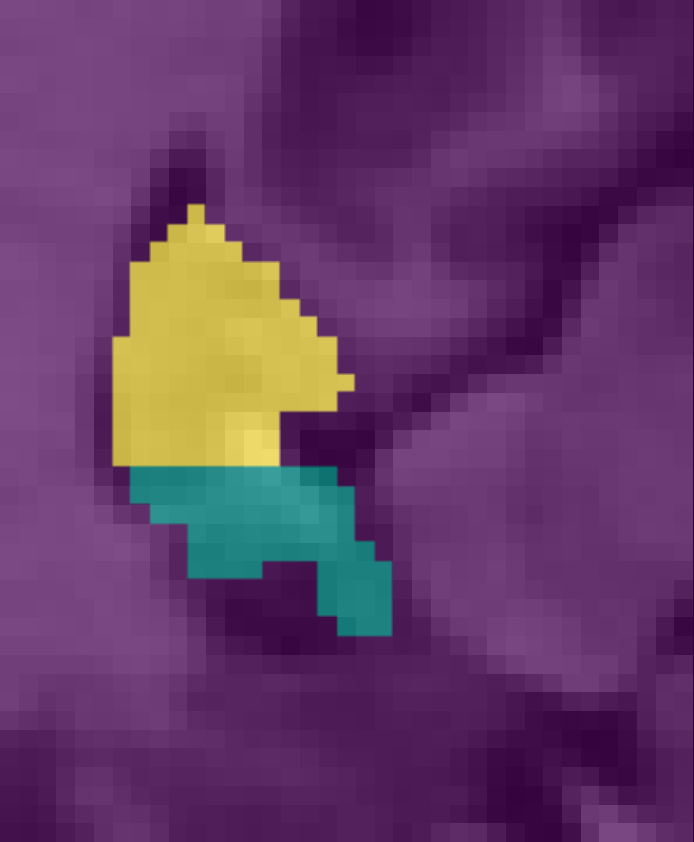}} &
    \multirow{1}{*}[25pt]{\revisiontwo{\shortstack[r]{w/o SB: \\ (PH\,/\,AH)} }} &
    \figSeg{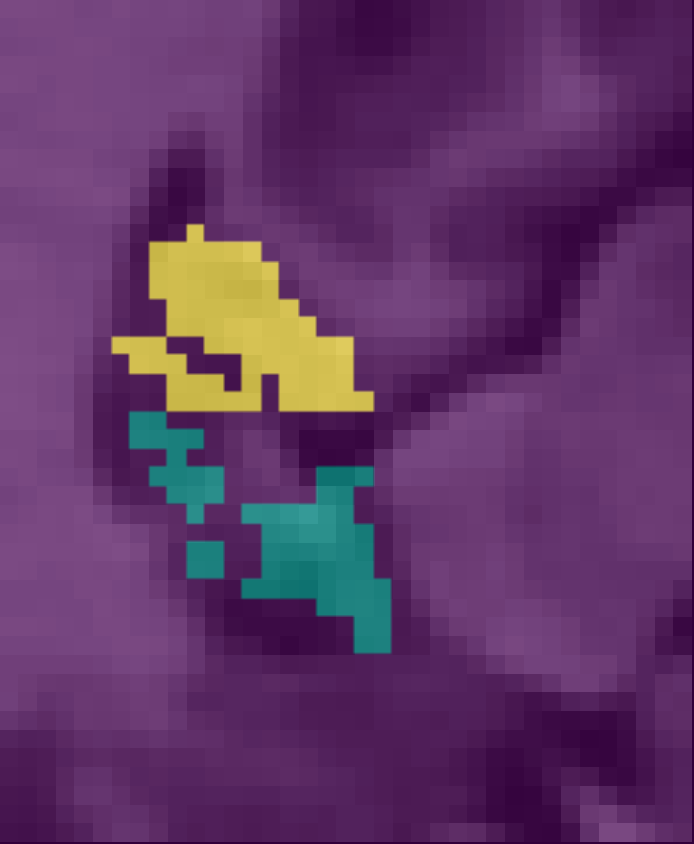} &
    \figSeg{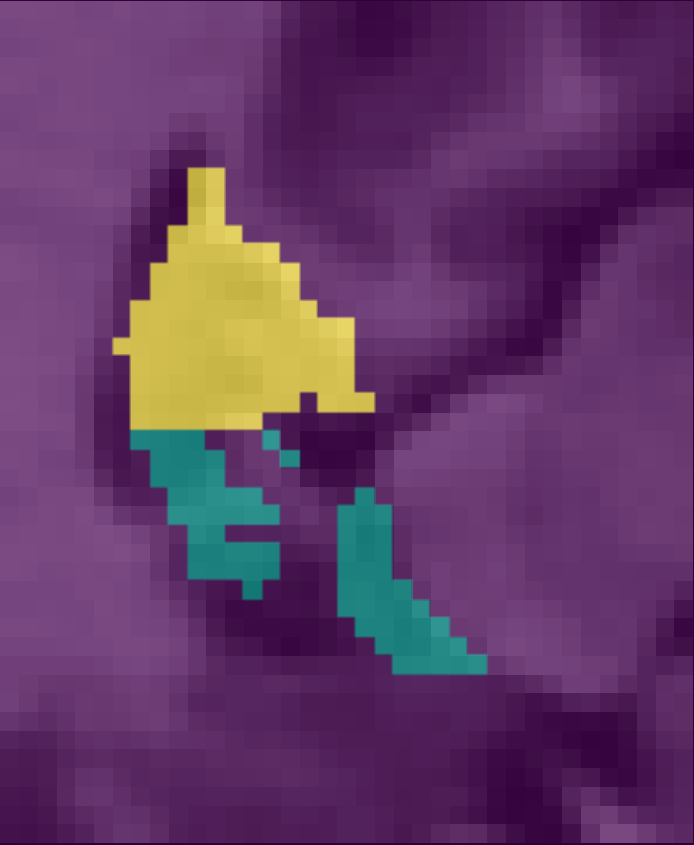} &
    \figSeg{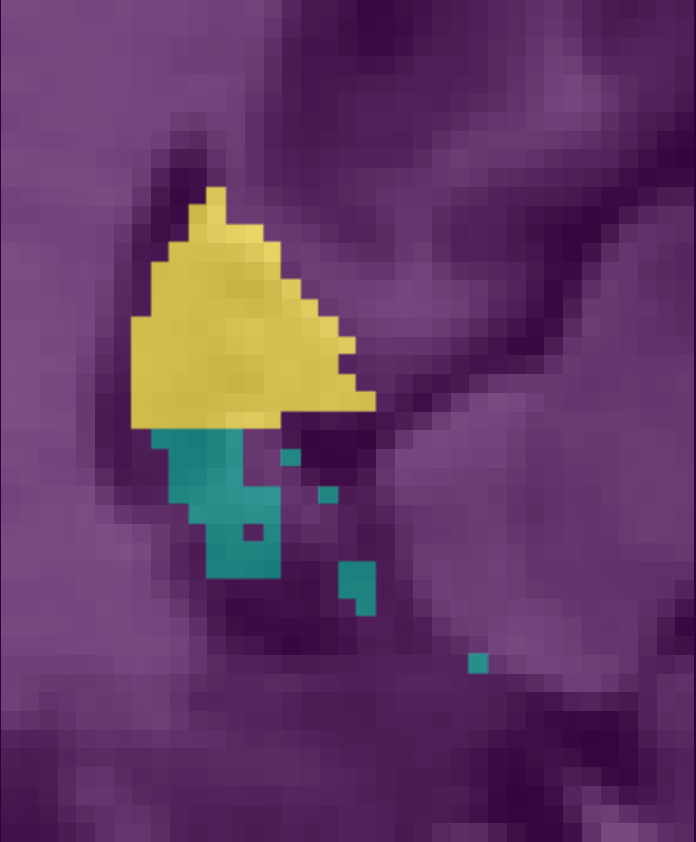} \\
     &  & \revisiontwo{71.13\,/\,66.18} & \revisiontwo{80.87\,/\,55.78} & \revisiontwo{82.14\,/\,53.91}\\[3pt]
    \revisiontwo{(a) Data \& Target}  & & \revisiontwo{(b) $2^{nd}$ cycle} & \revisiontwo{(c) $3^{rd}$ cycle} & \revisiontwo{(d) $4^{th}$ cycle}
    \end{tabular}
    \end{footnotesize}
    \caption{\textbf{Segmentation of a Hippocampus test sample across AL cycles}. The 2D dice score (DSC) is given for each predicted segmentation, both for the posterior hippocampus (PH, yellow) and the anterior hippocampus (AH, blue). At every AL cycle, the model trained on labelled samples selected with our stochastic batches (top row) predicts segmentations closer to the target mask (leftmost) compared to its purely Entropy-based counterpart (bottom row).
    }
    \label{fig:example_segmentation}
\end{figure}

We also visually investigate the benefits of using our stochastic batches with an uncertainty-based sample selection. In Fig.~\ref{fig:example_ALsamples}, we show two sets of candidate samples from the Prostate dataset identified by Entropy-based sampling, with and without our stochastic batches. \revisiontwo{The first two columns show samples} selected by identifying the most uncertain randomly generated batch. \revisiontwo{The last two columns depict the most certain queried samples} based on the individual entropy of their predicted output probabilities. While the samples from the first two columns seem more diverse, with more variety in the candidate set, the third column contains nearly identical samples. Indeed, tracking the first four images of the column to their corresponding 3D volume shows that the slices were taken from the MRI volume of the same patient. This confirms our claim that purely uncertainty-based strategies are likely to select very similar samples and that our stochastic batch sampling reduces the probability of querying samples with highly overlapping information.

Finally, we examine the impact of our selection strategy on the segmentation of test data. In Fig.~\ref{fig:example_segmentation}, we see that the model trained on images selected via our stochastic batch sampling method outputs better \revisiontwo{anterior and posterior hippocampus segmentations}. By the fourth cycle, the segmentation reaches a mean DSC \revisiontwo{(over both classes)} of 81.15\%, compared to the 68.03\% obtained via a purely Entropy-based sampling.

\subsection{Ablation experiments on the Prostate \revisiontwo{dataset}}

To evaluate the robustness of our method to different experimental settings, we perform a series of ablation studies on the Prostate dataset, evaluating the impact of the \revisiontwo{initial labelled set size, training hyper-parameters, sampling budget and stochastic pool size}.

\subsubsection{Impact of initial labelled set size}
\label{subsec:results_initlabeledsetsize}
For our first ablation study, we validate the performance of models trained on initial labelled sets of varying sizes. For each given initial labelled set size, the experiment is repeated with 5 initialization seeds controlling the initial labelled samples used, the model initialization and the training updates. Table \ref{table:results_numinitlabelled} gives the average model performance over 6 AL cycles. We observe that our stochastic batch selection strategy improves upon purely uncertainty-based selection \revisiontwo{also} when we vary the initial number of labelled samples.

\begin{table*}[bp]
\setlength{\tabcolsep}{4pt} 
\caption{\textbf{Overall improvements with Stochastic Batches for initial labelled sets of different sizes}. Mean model performance on the Prostate data over all AL cycles for initial sets of different sizes. We show the mean (std) Dice score (higher is better) over 3D test volumes \revisiontwo{(3D DSC)}. The results are averaged over 6 AL cycles (we omit results for the first AL cycle since all strategies share the same initial set). 
A * indicates the statistical significance of the result with a p-value $< 0.05$ given a paired permutation test.
}
{\small 
\begin{center}
\begin{tabular}{l c cc cc cc cc}
\toprule
 & \multirow[b]{3}{*}{RS} & \multicolumn{2}{c}{Entropy} & \multicolumn{2}{c}{Dropout} & \multicolumn{2}{c}{\revisiontwo{TTA}}  & \multicolumn{2}{c}{\revisiontwo{Learning Loss}}\\
 &  & \multicolumn{2}{c}{\scriptsize\citep{shannon_mathematical_1948}} & \multicolumn{2}{c}{\scriptsize\citep{gal_dropout_2016}} & \multicolumn{2}{c}{\scriptsize\citep{gaillochet_TAAL_2022}} & \multicolumn{2}{c}{\scriptsize\citep{yoo_learning_2019}} \\
 \cmidrule(l{5pt}r{5pt}){3-4}
 \cmidrule(l{5pt}r{5pt}){5-6}
 \cmidrule(l{5pt}r{5pt}){7-8}
 \cmidrule(l{5pt}r{5pt}){9-10}
 &   & w/o \textsc{SB} & Ours & w/o \textsc{SB} & Ours & w/o \textsc{SB} & Ours & w/o \textsc{SB} & Ours\\
\midrule
\multirow{2}{*}{\textbf{5 initial samples}} &  71.22 &  65.18 & \textbf{72.36*} &  61.69 & \textbf{71.45*} &  \revisiontwo{64.16} & \revisiontwo{\textbf{67.67}} & \revisiontwo{55.06} & \revisiontwo{\textbf{66.85*}}  \\
&  \ppm{15.09} &  \ppm{14.43} &  \ppm{16.19} &  \ppm{18.15} &  \ppm{15.41} &  \revisiontwo{\ppm{16.43}} &  \revisiontwo{\ppm{16.70}} &  \revisiontwo{\ppm{21.90}} &  \revisiontwo{\ppm{19.30}}  \\
\midrule 
\multirow{2}{*}{\textbf{10 initial samples}} &  71.08 &  66.21 & \textbf{73.78*} &  65.09 & \textbf{73.30*} &  \revisiontwo{70.23} & \revisiontwo{\textbf{73.01}} &  \revisiontwo{48.51} & \revisiontwo{\textbf{60.73*}}  \\
&  \ppm{11.70} &  \ppm{13.79} &  \ppm{10.73} &  \ppm{15.63} &  \ppm{14.95} &  \revisiontwo{\ppm{12.35}} &  \revisiontwo{\ppm{12.24}}  &  \revisiontwo{\ppm{9.83}} &  \revisiontwo{\ppm{9.10}}\\
\midrule 
\multirow{2}{*}{\textbf{15 initial samples}} &  75.21 &  0.7319 & \textbf{74.21} &  72.90 & \textbf{76.81*} &  \revisiontwo{\textbf{72.00}} & \revisiontwo{71.53} &  \revisiontwo{58.19} & \revisiontwo{\textbf{72.84*}} \\
&  \ppm{7.27} &  \ppm{7.54} &  \ppm{7.85} &  \ppm{6.96} &  \ppm{8.34} &  \revisiontwo{\ppm{10.86}} &  \revisiontwo{\ppm{8.84}} &  \revisiontwo{\ppm{12.09}} &  \revisiontwo{\ppm{10.32}} \\
\midrule 
\multirow{2}{*}{\textbf{20 initial samples}} &  76.00 &  76.13 & \textbf{80.24*} &  77.47 & \textbf{79.81*} &  \revisiontwo{74.59} & \revisiontwo{\textbf{78.04}} & \revisiontwo{69.81} & \revisiontwo{\textbf{75.51*}} \\
&  \ppm{7.19} &  \ppm{5.55} &  \ppm{4.19} &  \ppm{6.38} &  \ppm{05.54} &  \revisiontwo{\ppm{10.15}} &  \revisiontwo{\ppm{7.89}} &  \revisiontwo{\ppm{6.74}} &  \revisiontwo{\ppm{6.14}} \\
\midrule 
\multirow{2}{*}{\textbf{25 initial samples}} &  77.07 &  77.73 & \textbf{79.71*} &  77.44 & \textbf{81.08*}  &  \revisiontwo{76.81} & \revisiontwo{\textbf{78.32}} &  \revisiontwo{73.61} & \revisiontwo{\textbf{77.65*}} \\
&  \ppm{4.39} &  \ppm{3.79} &  \ppm{4.37} &  \ppm{4.31} &  \ppm{5.20} &  \revisiontwo{\ppm{9.03}} &  \revisiontwo{\ppm{5.88}} &  \revisiontwo{\ppm{5.27}} &  \revisiontwo{\ppm{5.52}}  \\
\bottomrule
\end{tabular}
\end{center}
}
\label{table:results_numinitlabelled}
\end{table*}

\subsubsection{Impact of training hyper-parameters}
\label{subsec:results_hyperparams}

\begin{figure}[htb]
    \centering
    \begin{subfigure}{0.5\textwidth}
        \includegraphics[width=\linewidth]{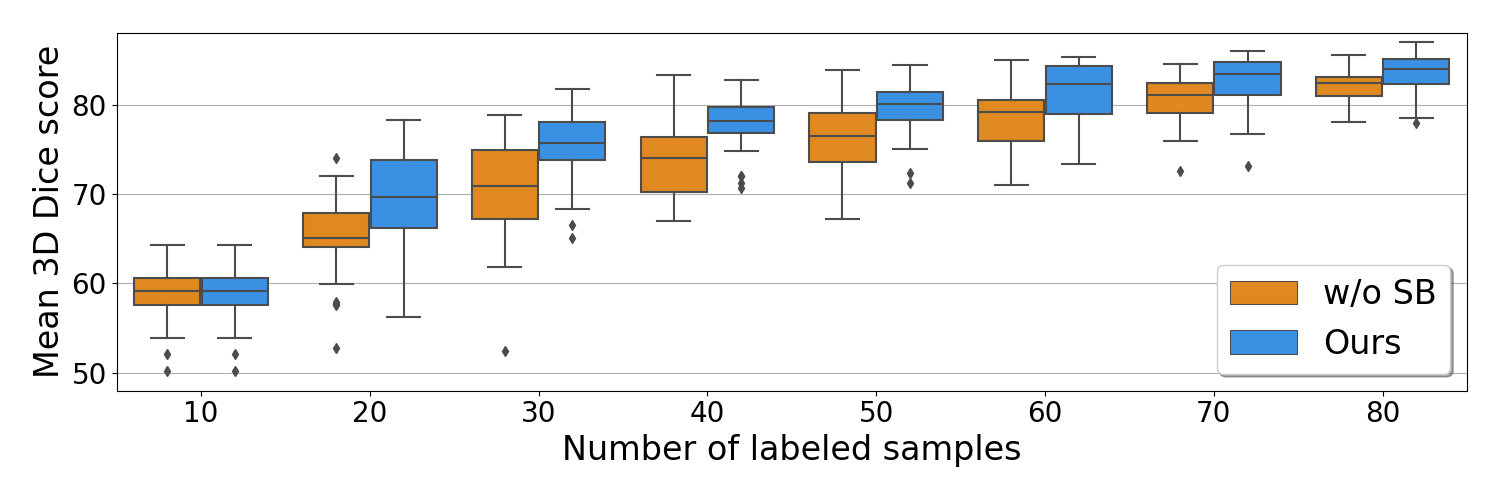}
        \caption{Improvements for Entropy \citep{shannon_mathematical_1948} \label{fig:sub:results_hyperparams_entropy}}
    \end{subfigure}\hfil 
    \hfil
    \begin{subfigure}{0.5\textwidth}
        \includegraphics[width=\linewidth]{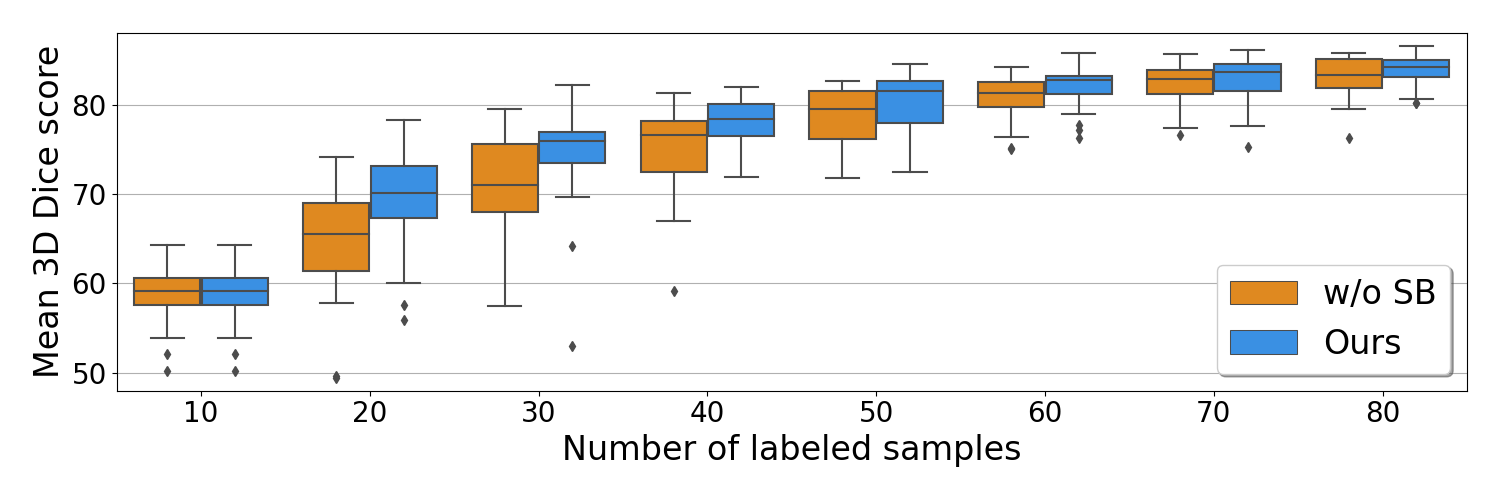}
        \caption{Improvements for Dropout \citep{gal_dropout_2016} \label{fig:sub:results_hyperparams_dropout}}
    \end{subfigure}\hfil 
    \medskip 
    \begin{subfigure}{0.5\textwidth}
        \includegraphics[width=\linewidth]{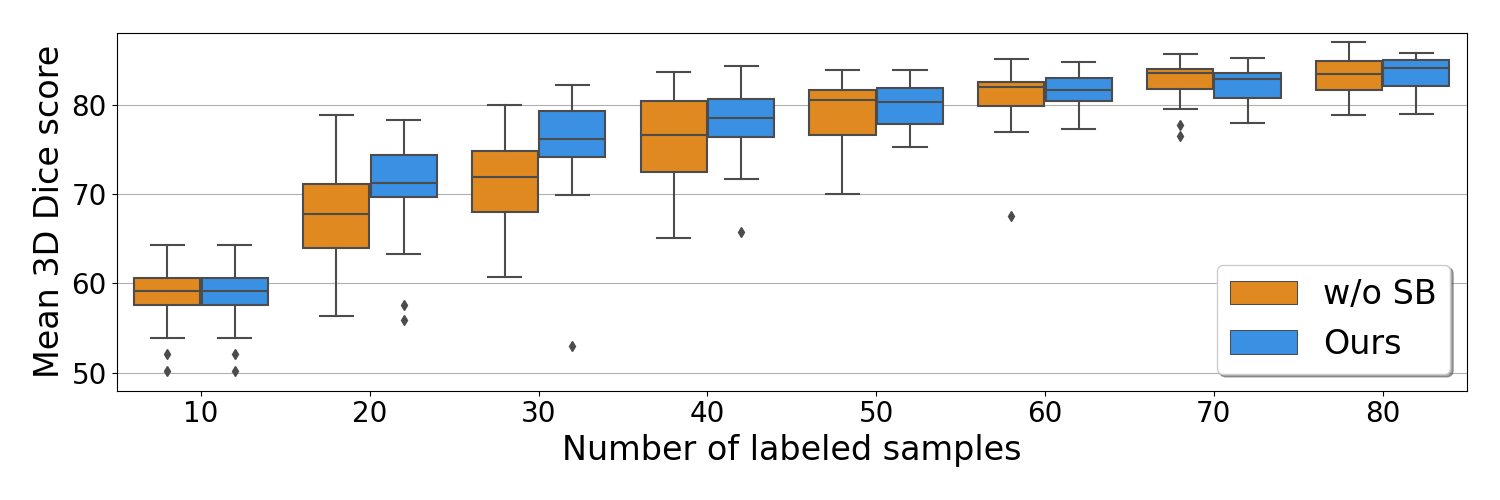}
        \caption{Improvements for TTA \citep{gaillochet_TAAL_2022} \label{fig:sub:results_hyperparams_TTA}}
    \end{subfigure}\hfil 
    \hfil
    \begin{subfigure}{0.5\textwidth}
        \includegraphics[width=\linewidth]{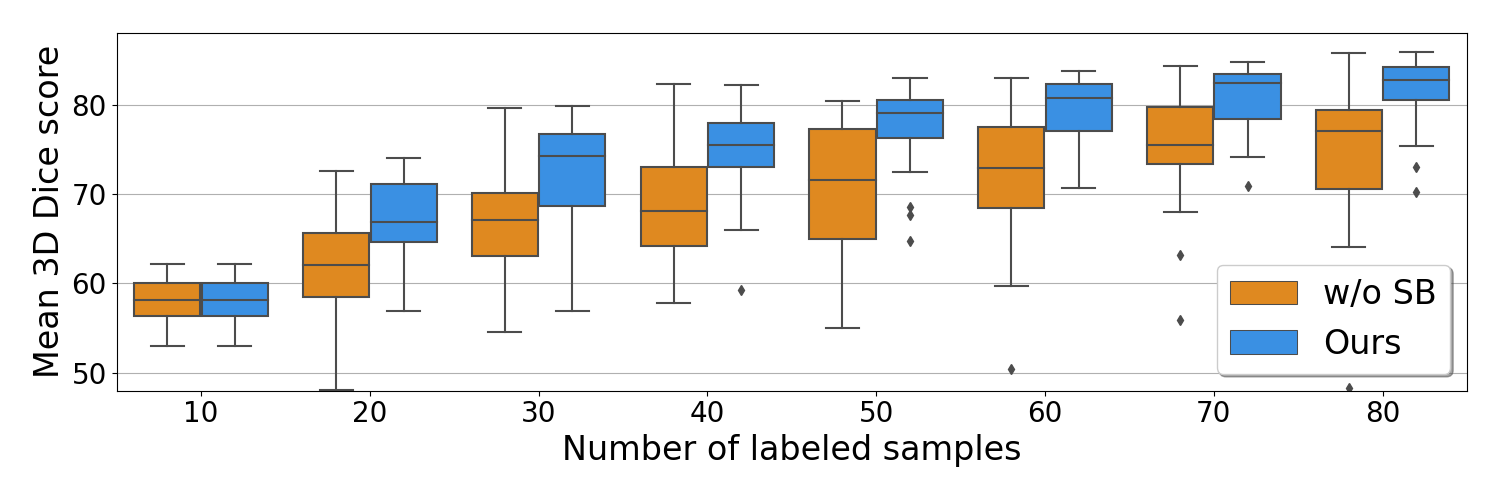}
        \caption{Improvements for Learning loss \citep{yoo_learning_2019} \label{fig:sub:results_hyperparams_learningloss}}
    \end{subfigure}\hfil 
    \caption{\textbf{Improvements with Stochastic Batches over varying hyper-parameters}. \revisiontwo{Box plot} of active learning results on \revisiontwo{Prostate data} in terms of 3D test dice score, given over 5 training hyper-parameters sets and 5 initialization seeds. Depicted are the results for sampling based on \subref{fig:sub:results_hyperparams_entropy}) Entropy, \subref{fig:sub:results_hyperparams_dropout}) Dropout, 
    \subref{fig:sub:results_hyperparams_TTA}) Test-time augmentation and \subref{fig:sub:results_hyperparams_learningloss}) Learning loss. The AL selection is shown with (blue) and without (orange) stochastic batches. Our stochastic batches improve the model performance of purely uncertainty-based AL strategies and boost performance, even with variations in hyper-parameters.
    }
    \label{fig:results_hyperparams}
\end{figure}

\begin{table*}[htb]
\setlength{\tabcolsep}{4pt} 
\caption{\textbf{Overall improvements with Stochastic Batches over varying training hyper-parameters}. Mean model performance on \revisiontwo{Prostate data} over all AL cycles (omitting training with the initial labelled set). We show the mean (std) Dice score (DSC, higher is better) and $95\%$ Hausdorff (HD95, lower is better) distance 
over 3D test volumes and individual 2D test images. The results are averaged over 7 AL cycles and 5 training hyper-parameter sets. * indicates the statistical significance of the result with a p-value $< 0.05$ given a paired permutation test.
}
{\small \begin{center}
\begin{tabular}{l c cc cc cc cc}
\toprule
 & \multirow[b]{3}{*}{RS} & \multicolumn{2}{c}{Entropy} & \multicolumn{2}{c}{Dropout} & \multicolumn{2}{c}{TTA} & \multicolumn{2}{c}{Learning Loss}\\
 &  & \multicolumn{2}{c}{\scriptsize\citep{shannon_mathematical_1948}} & \multicolumn{2}{c}{\scriptsize\citep{gal_dropout_2016}} & \multicolumn{2}{c}{\scriptsize\citep{gaillochet_TAAL_2022}} & \multicolumn{2}{c}{\scriptsize\citep{yoo_learning_2019}}\\
 \cmidrule(l{5pt}r{5pt}){3-4}
 \cmidrule(l{5pt}r{5pt}){5-6}
 \cmidrule(l{5pt}r{5pt}){7-8}
 \cmidrule(l{5pt}r{5pt}){9-10}
&   & w/o \textsc{SB} & Ours & w/o \textsc{SB} & Ours & w/o \textsc{SB} & Ours & w/o \textsc{SB} & Ours\\
\midrule
\multirow{2}{*}{\textbf{3D DSC} ($\uparrow$\,best)} & 75.57 &  75.13 &  \textbf{78.44*} &  76.49 &  \textbf{78.59*} &  77.33 &  \textbf{78.67*}  &  69.53 &  \textbf{76.25*}\\
 & \ppm{6.48} &  \ppm{6.95} &  \ppm{6.02} &  \ppm{7.65} &  \ppm{6.09}  &  \ppm{6.92} &  \ppm{5.53} &  \ppm{8.43} &  \ppm{6.68}\\
\midrule
\multirow{2}{*}{\textbf{2D DSC} ($\uparrow$\,best)}&  68.29 &  68.90 &  \textbf{71.04*} &  69.62 &  \textbf{71.08*} &  70.46 &  \textbf{71.31*}  &  64.27 &  \textbf{69.16*}\\
&  \ppm{6.79} &  \ppm{7.34} &  \ppm{6.51} &  \ppm{6.70} &  \ppm{6.79} &  \ppm{7.05} & \ppm{5.71}  &  \ppm{7.23} &  \ppm{6.80}\\
\midrule
\multirow{2}{*}{\textbf{3D HD95} ($\downarrow$\,best)} &  7.58 &  7.87 &  \textbf{6.83*} &  \textbf{6.72} &  6.74 &  6.32 &  \textbf{6.13} &  8.78 &  \textbf{7.85*}\\
&  \ppm{3.86} &  \ppm{4.28} &  \ppm{3.31} &  \ppm{2.75} &  \ppm{3.29} &  \ppm{2.87} & \ppm{2.82}  &  \ppm{4.22} &  \ppm{3.68}\\
\bottomrule
\end{tabular}
\end{center}
}
\label{table:results_hyperparams}
\end{table*}

Active Learning methods typically tune hyper-parameters using an initial labelled set, maintaining these settings throughout all AL cycles. However, these parameters might be sub-optimal for subsequent training cycles as more labelled data becomes available. We hence explore the robustness of stochastic batches to different yet realistic training hyperparameters. We select five hyperparameter sets, each optimized for labelled set sizes of 10, 50, 100, 150, and 200. These sets included diverse augmentation parameters, scheduling parameters and loss function weights.

Results in Fig.~\ref{fig:results_hyperparams} reveal that our stochastic batch sampling noticeably improves the performance of purely uncertainty-based sampling, particularly in the first 3 or 4 AL cycles. In addition, the spread of 3D dice scores tends to be narrower with our method than with a purely uncertainty-based sampling, showing that our strategy tends to be more stable.

The benefit of using our stochastic batches is most evident in the average dice scores over all AL cycles for both test images and volumes, as given in Tab.~\ref{table:results_hyperparams}. Test-Time Augmentation (TTA) generally performs better with stochastic batches, although the results are not statistically significant for distance-based metrics. This could be due to the fact that we vary the training and regularization hyper-parameters while keeping data augmentation parameters fixed for sampling.

\subsubsection{Impact of sampling budget}
\label{subsec:results_budget}

\begin{figure}[t!] 
   \centering
   \begin{subfigure}{0.49\textwidth}
       \includegraphics[width=\linewidth]{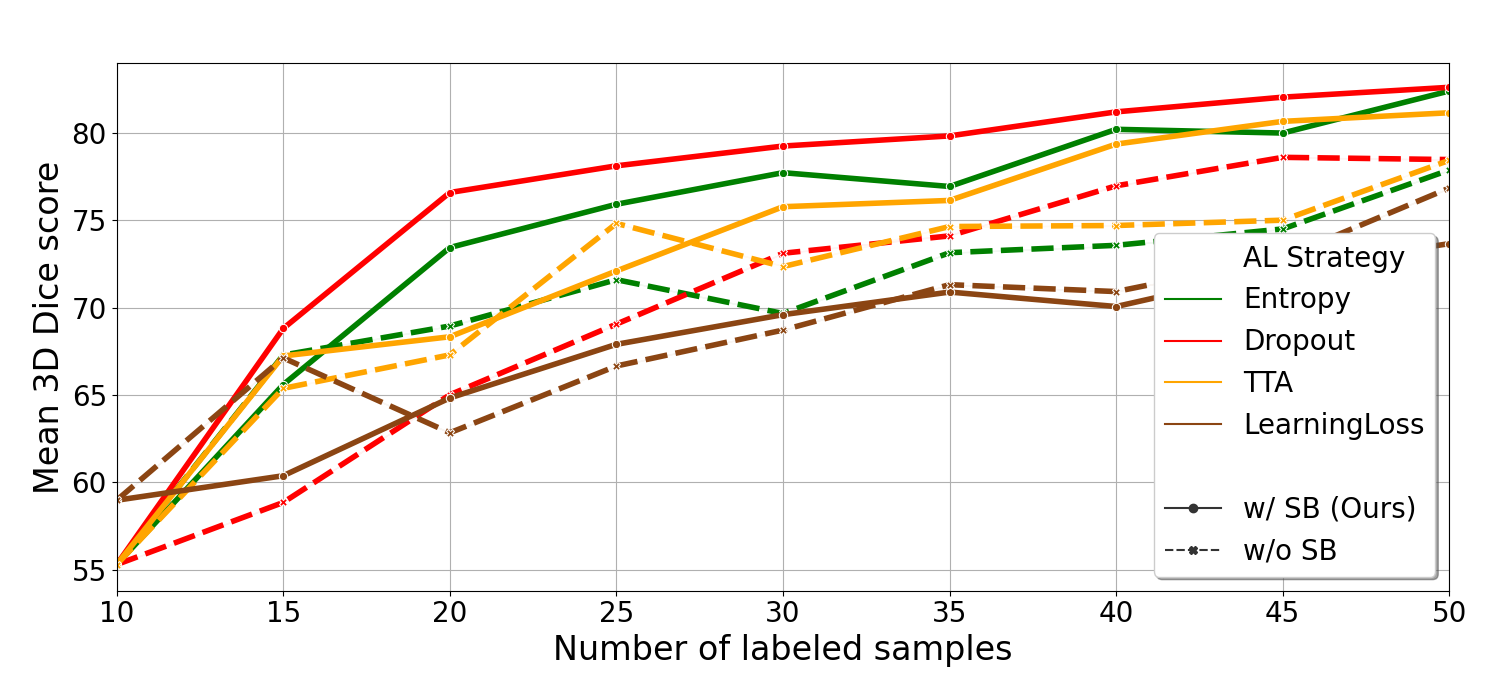}
       \caption{Improvements with low budget ($B = 5$)}
       \label{fig:sub:results_budget5}
   \end{subfigure}\hfil 
   \hfil
    \begin{subfigure}{0.49\textwidth}
       \includegraphics[width=\linewidth]{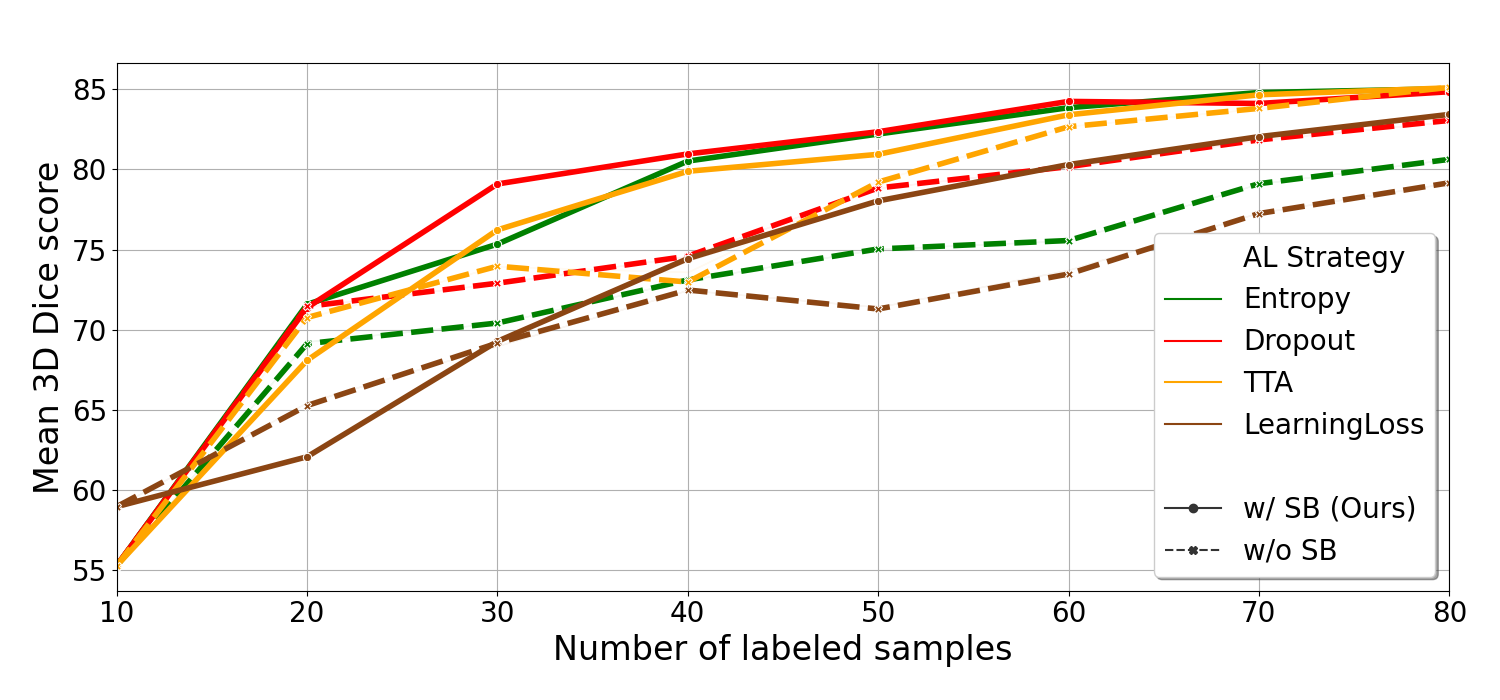}
       \caption{Improvements with mid budget ($B = 10$)}
       \label{fig:sub:results_budget10}
   \end{subfigure}\hfil 
    \hfil
   \begin{subfigure}{0.49\textwidth}
       \includegraphics[width=\linewidth]{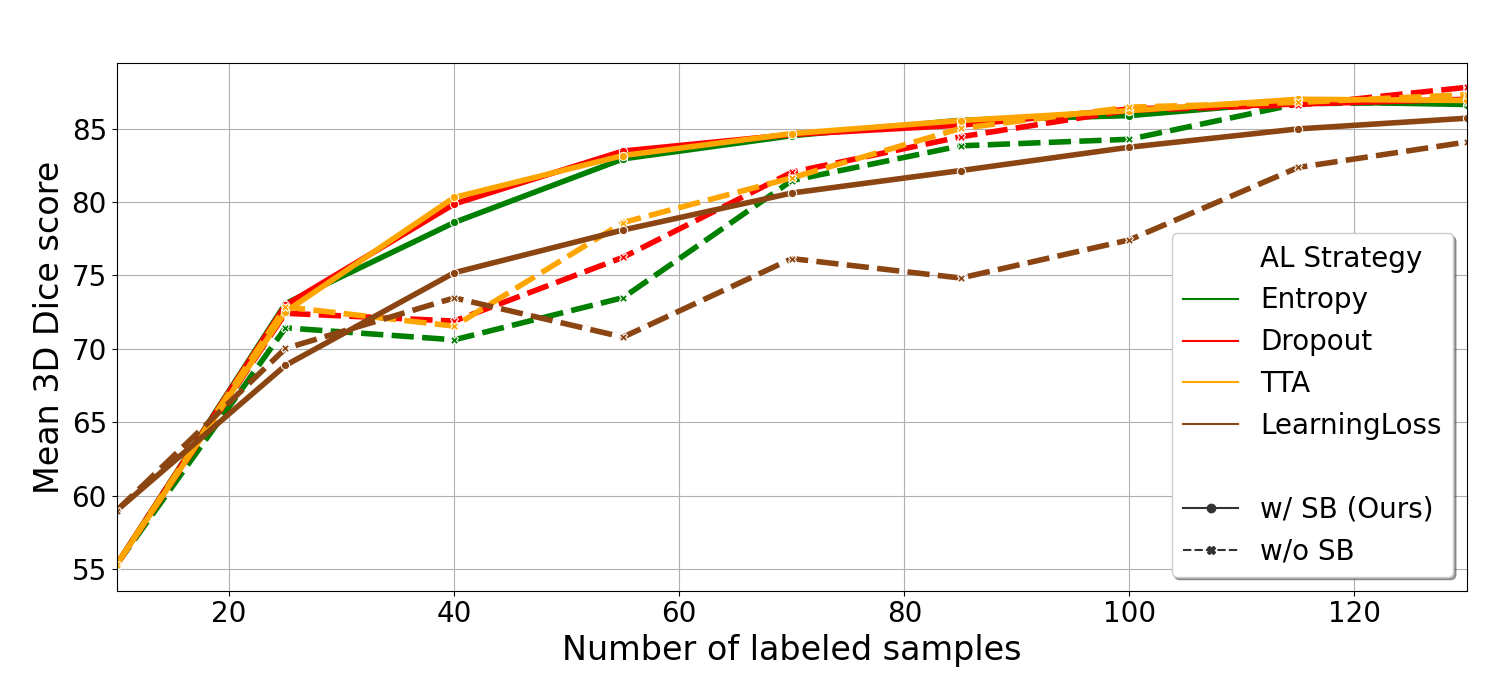}
       \caption{Improvements with high budget ($B = 15$)}
       \label{fig:sub:results_budget15}
   \end{subfigure}\hfil 
   \caption{\revisiontwo{\textbf{Improvements with Stochastic Batches given different budget sizes}}. Model performance in terms of 3D dice score on test volumes given active learning selection with (solid) and without (dashed) stochastic batches on Prostate data. \revisiontwo{The results are given for sampling budgets \subref{fig:sub:results_budget5}) $B\! =\! 5$, \subref{fig:sub:results_budget10}) $B\! =\! 10$ and \subref{fig:sub:results_budget15}) $B\! =\! 15$. Depicted are the results for sampling based on Entropy (green) and Dropout (red), TTA (yellow) and Learning Loss (brown).}
   Using stochastic batches during sampling improves the model performance at both low and higher budgets.}
   \label{fig:results_budget}
\end{figure}

We also investigate the robustness of stochastic batches to the sampling budget. Keeping the initial labelled set and training hyper-parameters fixed, we run experiments with 5 different sampling budgets, which we keep constant across cycles. In this experiment, since we vary $B$, images are allowed to be resampled when generating the stochastic batches, and we keep the number of generated batches to a fixed $Q=100$.

The results shown in Fig.~\ref{fig:results_budget} reveal that stochastic batches have a more consistent impact on model performance as the budget size increases. 
With a high budget $B=15$, the use of stochastic batches constantly improves purely uncertainty-based methods. An improvement is also visible for lower budgets, such as $B=5$, particularly for the Entropy, Dropout and TTA-based sampling. 

However, with very low budgets, batch uncertainty is highly influenced by the uncertainty of each individual sample, potentially reducing the benefits of diversity offered by stochastic batches. The selection is dominated by uncertainty, and if the measure for uncertainty is not representative of the true uncertainty of the model, then uninformative samples could be selected and consequently bias the model.

\subsubsection{Impact of sampling stochastic pool size}
\label{subsec:results_sizepool}
\begin{figure}[t]
    \centering
    \begin{subfigure}{0.5\textwidth}
        \includegraphics[width=\linewidth]{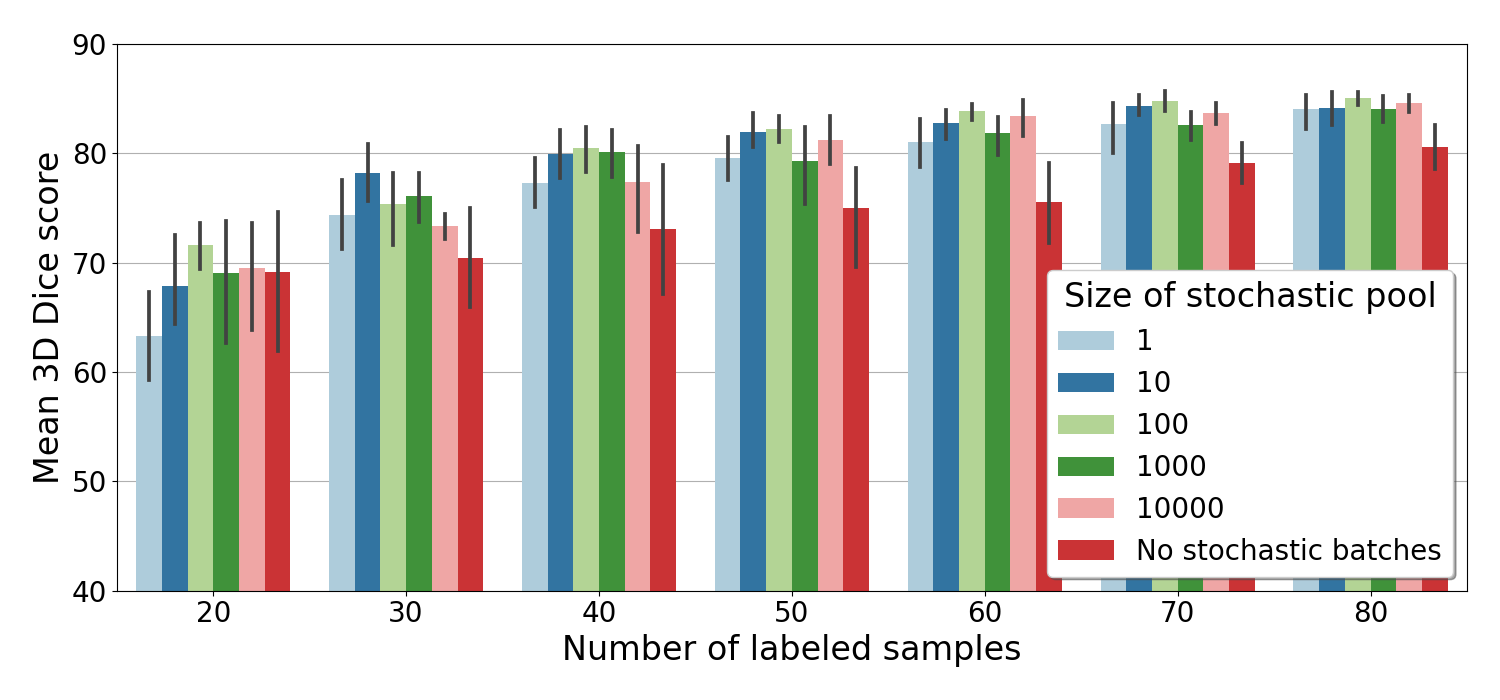}
        \caption{Stochastic batches for Entropy \citep{shannon_mathematical_1948} \label{fig:sub:results_numpacks_entropy}}
    \end{subfigure}\hfil 
    \hfil
    \begin{subfigure}{0.5\textwidth}
      \includegraphics[width=\linewidth]{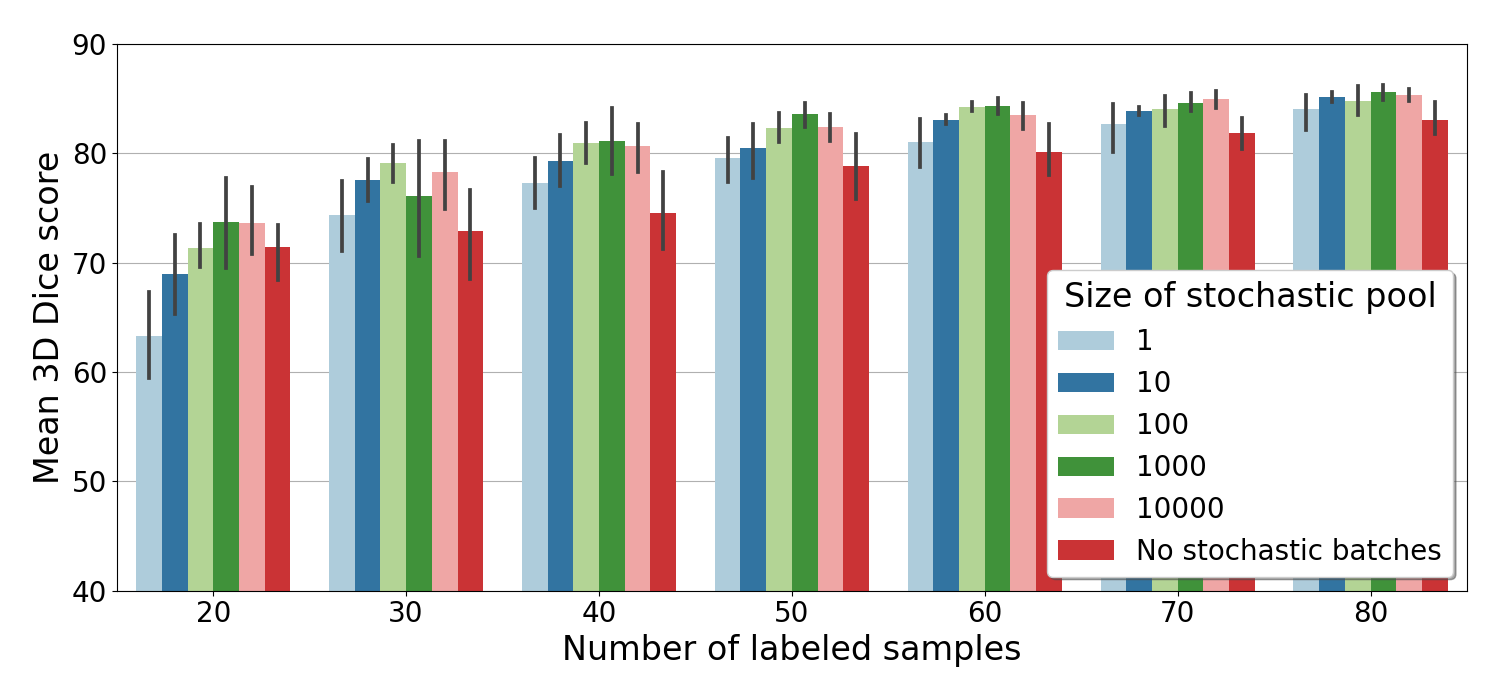}
      \caption{Stochastic batches for Dropout \citep{gal_dropout_2016}\label{fig:sub:results_numpacks_dropout}}
    \end{subfigure}\hfil 
    \medskip 
    \caption{\textbf{Impact of pool size of Stochastic Batches}. Model performance in terms of 3D dice score on test volumes from \revisiontwo{Prostate data} given stochastic batch pools of different sizes. The error bars (black) corresponds to the $95\%$ confidence interval over 5 experiments with different seed initialization. Depicted are the results 2 popular uncertainty-based AL methods: Entropy-based sampling (\ref{fig:sub:results_numpacks_entropy}) and Dropout-based sampling (\ref{fig:sub:results_numpacks_dropout}). 
    A medium pool size between 10 to 100 yields some of the most advantageous performances.}
    \label{fig:results_numpacks}
\end{figure}

In our last ablation study, we evaluate the influence of the number of batches in the stochastic pool on the model performance, fixing the initial labelled set, training hyper-parameters and sampling budget. Instead of generating $Q = \mathrm{floor}(\nicefrac{|\mathcal{D}_u|}{B})$ batches, we artificially vary $Q$. Accordingly, we allow resampling so samples can appear in multiple generated batches. 
The results for our experiments on Entropy-based and Dropout-based sampling are given in Fig.~\ref{fig:results_numpacks}. Applying the biggest pool size does not necessarily yield the best performance. On the contrary, the model performs best when the most uncertain batch is selected from a pool containing 10 or 100 different batches. Increasing the pool of choices by 10 or 100 does not lead to significant improvements and can lead to worse performances. 

\section{Discussion}
\label{sec:discussion}

Overall, our results demonstrate that using stochastic batches during uncertainty-based sampling is an efficient strategy to ensure diversity among the selected batch of samples. Furthermore, we experimentally observe that the benefit of using stochastic batches is robust to changes in the initial labelled set, initialization of the model and training hyper-parameters, as well as to variations in the sampling budget. 

As illustrated in Fig.~\ref{fig:example_ALsamples}, the redundancy of queried samples constitutes one of the main drawbacks of uncertainty-based AL strategies. Their queried samples may indeed convey highly similar information. Hence, the annotation effort on these samples will be suboptimal. If, on the contrary, the most uncertain batches rather than the most uncertain samples are queried, the added diversity within our stochastic batches mitigates the overlap of information and redundancy between samples. Our stochastic scheme adds diversity to the uncertainty-based sampling in AL in a fast, computationally-efficient way, as shown by Tab.~\ref{table:results_time}. Our quantitative results demonstrate the advantages of adding such a stochastic scheme in AL in terms of added segmentation accuracy in a low-labelled set regime and reduced number of required training samples.

Previous AL works have observed that the initial labelled pool can significantly impact the training and final performance of AL models \citep{chen_making_2022}. Nevertheless, a robust AL method should still perform well regardless of this initial labelled set. The results obtained in our experiment with varying initial labelled sets (Sec.~\ref{subsec:results_overall} and Sec.~\ref{subsec:results_initlabeledsetsize}) reveal that the performance boost from our stochastic batch sampling strategy is robust to changes in both the initial labelled set and model initialization. On average, selecting the most uncertain batches across AL cycles yields better results than selecting the most uncertain samples.
Similarly, Sec.~\ref{subsec:results_hyperparams} shows that the improvements yielded by stochastic AL batches are also robust to changes in the training and regularization parameters. Hence, our method can maintain efficiency despite changes in the learning environment. These results suggest that using stochastic batches during AL for uncertainty-based sampling can be a reliable and robust AL approach. 

Our stochastic batch querying strategy for uncertainty-based AL operates as a balance between a fully random and a purely uncertainty-based selection. While we set $Q = \mathrm{floor}\big(\nicefrac{|\mathcal{D}_U|}{B})$, the stochastic pool size $Q$ can also be directly modified to control the amount of randomness desired in the AL selection.
With the smallest pool size ($Q = 1$), our stochastic batch selection is equivalent to random sampling since the single suggested batch will automatically have the highest uncertainty score in the pool. With the biggest pool size ($Q \rightarrow \infty$), all possible combinations of samples are available in the pool, and selecting the most uncertain batch of samples is equivalent to selecting the top uncertain samples. In other words, the approach becomes a purely uncertainty-based AL strategy with a larger pool size. 
As shown in Sec.~\ref{subsec:results_sizepool}, the benefits of our stochastic batches are apparent in between those extreme $Q$ values, when the sampling strategy combines the informativeness of uncertainty-based sampling with the diversity provided by random sampling. 
Active learning is an expensive framework to experiment with, given that AL cycles are iterative and that procedures should be repeated to reduce as much as possible the influence of initialization. In this work, we ran multiple experiments with different settings (size and type of initial labelled set, training hyper-parameters, stochastic pool size, sampling budget) to test how stable our method was. However, we acknowledge that \revisiontwo{our experiments do not cover} all ranges of possible setups.

\section{Conclusion}
\label{sec:conclusion}
Active learning is particularly relevant in medical image segmentation since manual labelling is highly time-consuming and expensive. This paper addresses three main limitations of AL strategies: the \revisiontwo{relatively} limited literature on AL work for medical image segmentation compared to classification tasks, the tendency of uncertainty-based batch sampling strategies to select very similar samples and the computational burden of diversity-based methods. Instead of employing sample-level uncertainty for candidate selection, we suggest a batch-level approach where uncertainty is computed over randomly generated batches of samples.
Using stochastic batches with uncertainty-based sampling is a simple, computational-inexpensive approach to improve the AL candidate selection and, hence, the final model performance. Our method is flexible and easily adaptable to any uncertainty-based AL strategy. 
In addition, \revisiontwo{our extensive experiments} show that adding stochastic batches improves purely uncertainty-based methods consistently across different experimental setups. \revisiontwo{Hence, stochastic batching could bring a more reliable advantage over other representative-based works, which have shown significantly varying amounts of robustness in performance \citep{munjal_towards_2022}.} Our method could therefore act as a strong baseline to better use the limited annotation time of clinical experts when segmenting medical images.

\section*{Acknowledgments}

This work is supported by the Canada Research Chair on Shape Analysis in Medical Imaging, the Research Council of Canada (NSERC) and the Quebec Bio-Imaging Network (QBIN). Computational resources were partially provided by Compute Canada. The authors also thank the PROMISE12 and the Medical Segmentation Decathlon challenge organizers for providing the data.

\bibliographystyle{unsrtnat}
\bibliography{references_MedIA}  

\end{document}